\documentclass{article}

\usepackage{microtype}
\usepackage{graphicx}
\usepackage{subfigure}
\usepackage{booktabs} 

\usepackage{hyperref}

\usepackage[accepted]{icml2024}

\usepackage{amsmath}
\usepackage{amssymb}
\usepackage{mathtools}
\usepackage{amsthm}
\usepackage{multirow}
\usepackage{multicol}
\usepackage{colortbl}
\usepackage{bbding}
\definecolor{mygray}{RGB}{230, 230, 230}
\definecolor{myred}{RGB}{250, 127, 111}

\usepackage[capitalize,noabbrev]{cleveref}

\theoremstyle{plain}
\newtheorem{theorem}{Theorem}[section]

\newtheorem{lemma}[theorem]{Lemma}

\theoremstyle{definition}

\theoremstyle{remark}

\usepackage[textsize=tiny]{todonotes}

\icmltitlerunning{A RG Flow-Based Regularization for GAN Training with Limited Data}

\begin{document}

\twocolumn[
\icmltitle{MS$^3$D: A RG Flow-Based Regularization for GAN Training with Limited Data}



\icmlsetsymbol{equal}{*}

\begin{icmlauthorlist}
\icmlauthor{Jian Wang}{yyy}
\icmlauthor{Xin Lan}{yyy}
\icmlauthor{Yuxin Tian}{yyy}
\icmlauthor{Jiancheng Lv}{yyy}
\end{icmlauthorlist}

\icmlaffiliation{yyy}{College of Computer Science,
Sichuan University and Engineering Research Center of Machine Learning and Industry Intelligence, Ministry of Education, Chengdu 610065, P.R. China}

\icmlcorrespondingauthor{Jiancheng Lv}{jianwang.scu@gmail.com;lvjiancheng@scu.edu.cn}

\icmlkeywords{GAN, Limited data, RG flow}

\vskip 0.3in
]



\printAffiliationsAndNotice{} 

\begin{abstract}
Generative adversarial networks (GANs) have made impressive advances in image generation, but they often require large-scale training data to avoid degradation caused by discriminator overfitting. To tackle this issue, we investigate the challenge of training GANs with limited data, and propose a novel regularization method based on the idea of renormalization group (RG) in physics.We observe that in the limited data setting, the gradient pattern that the generator obtains from the discriminator becomes more aggregated over time. In RG context, this aggregated pattern exhibits a high discrepancy from its coarse-grained versions, which implies a high-capacity and sensitive system, prone to overfitting and collapse. To address this problem, we introduce a \textbf{m}ulti-\textbf{s}cale \textbf{s}tructural \textbf{s}elf-\textbf{d}issimilarity (MS$^3$D) regularization, which constrains the gradient field to have a consistent pattern across different scales, thereby fostering a more redundant and robust system. We show that our method can effectively enhance the performance and stability of GANs under limited data scenarios, and even allow them to generate high-quality images with very few data.
\end{abstract}

\section{Introduction}
The challenge of training GANs with limited data has garnered increasing attention within the research community. Numerous studies~\cite{nips/ZhaoLLZ020,nips/KarrasAHLLA20,nips/JiangDWL21,cvpr/TsengJL0Y21,nips/Fang0S22,cvpr/CuiYZLLX23} have reached a consensus: insufficient data often leads to overfitting in the discriminator. This overfitting results in a lack of meaningful dynamic guidance for the generator, ultimately causing its performance to degrade. Data augmentation, a standard solution to prevent overfitting in deep learning~\cite{jbd/ShortenK19}, has been widely applied to GAN training in limited data scenarios. Traditional~\cite{corr/abs-2006-02595,tip/TranTNNC21} and differentiable~\cite{nips/ZhaoLLZ020} data augmentation techniques for both real and generated images, as well as various adaptive augmentation strategies~\cite{nips/KarrasAHLLA20}, have shown good performance on several standard benchmarks. However, the effectiveness of data augmentation heavily depends on specific handcrafted types or a costly search process, which limits its generality~\cite{corr/abs-2006-02595,nips/KarrasAHLLA20}. Additionally, data augmentation addresses data deficiency by enhancing the quantity and diversity of samples without providing a deeper understanding of the internal dynamics of GANs.

\begin{figure}[!tbp]
\centering
\setlength{\tabcolsep}{0pt}
\subfigure[RG flow: fine-grained $\to$ coarse-grained.]{
\centering
{\small
\centering
\begin{tabular}{c c c c c c c c}
\multicolumn{8}{c}{Training step: 1.2M, $\mathcal{D}_{\Gamma}=0.0825$}
\tabularnewline
\includegraphics[width=0.122\linewidth]{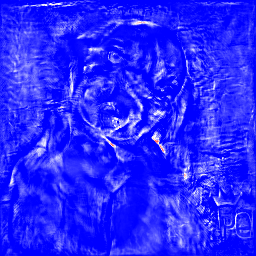}&
\includegraphics[width=0.122\linewidth]{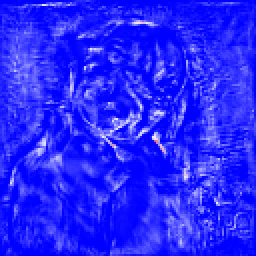}&
\includegraphics[width=0.122\linewidth]{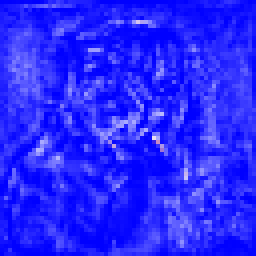}&
\includegraphics[width=0.122\linewidth]{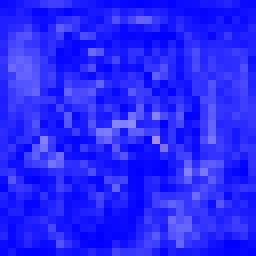}&
\includegraphics[width=0.122\linewidth]{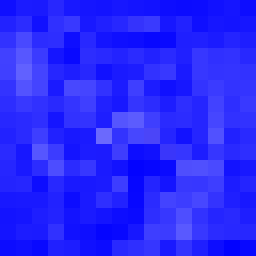}&
\includegraphics[width=0.122\linewidth]{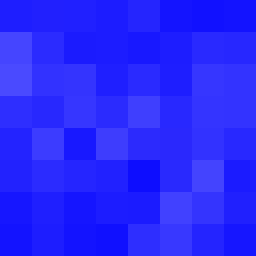}&
\includegraphics[width=0.122\linewidth]{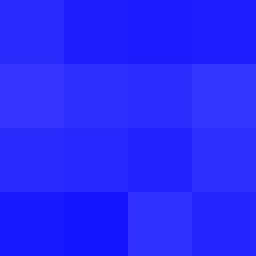}&
\includegraphics[width=0.122\linewidth]{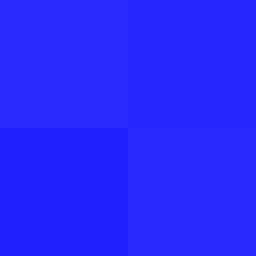}
\tabularnewline
\multicolumn{8}{c}{Training step: 5M, $\mathcal{D}_{\Gamma}=0.0872$}
\tabularnewline
\includegraphics[width=0.122\linewidth]{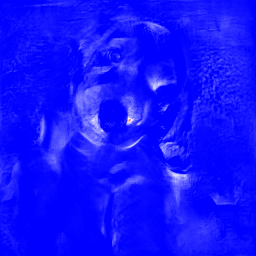}&
\includegraphics[width=0.122\linewidth]{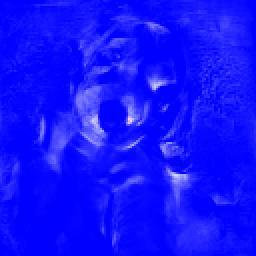}&
\includegraphics[width=0.122\linewidth]{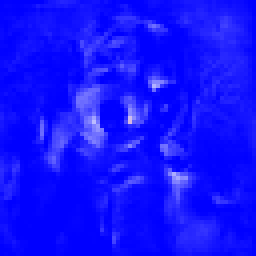}&
\includegraphics[width=0.122\linewidth]{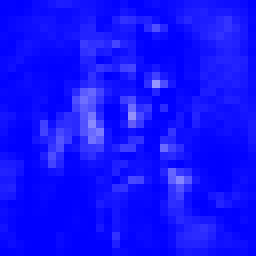}&
\includegraphics[width=0.122\linewidth]{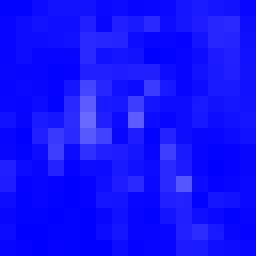}&
\includegraphics[width=0.122\linewidth]{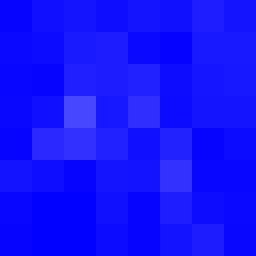}&
\includegraphics[width=0.122\linewidth]{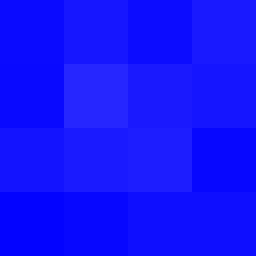}&
\includegraphics[width=0.122\linewidth]{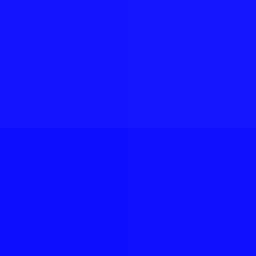}
\end{tabular}
}
}
\vspace{-0.02in}
\vspace{-0.08in}
\subfigure[Snapshots of food coloring diffusion in hot water over time.]
{{\small
\centering
\setlength{\tabcolsep}{0pt}
\begin{tabular}{c c c c c}
$\mathcal{D}_{\Gamma}=$0.0713 & 0.0658 & 0.0632 & 0.0623 & 0.0567
\tabularnewline
\includegraphics[width=0.1975\linewidth]{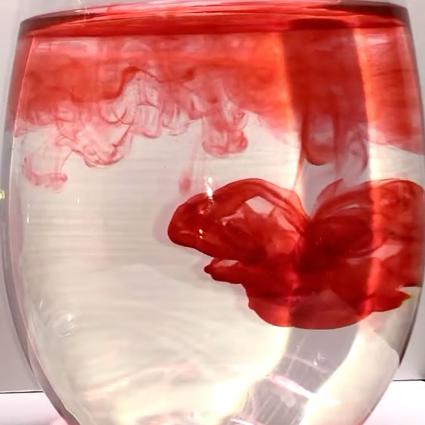}&
\includegraphics[width=0.1975\linewidth]{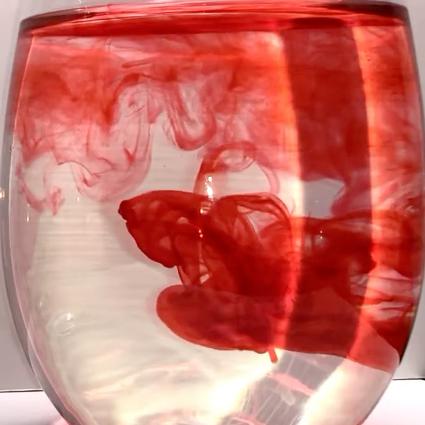}&
\includegraphics[width=0.1975\linewidth]{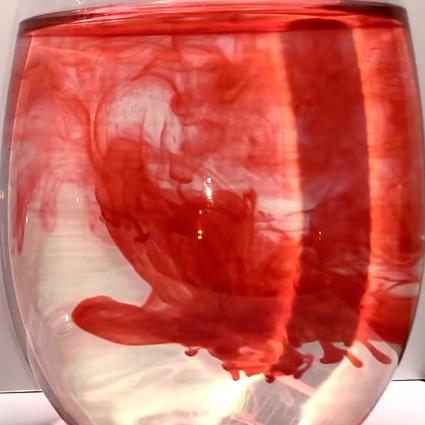}&
\includegraphics[width=0.1975\linewidth]{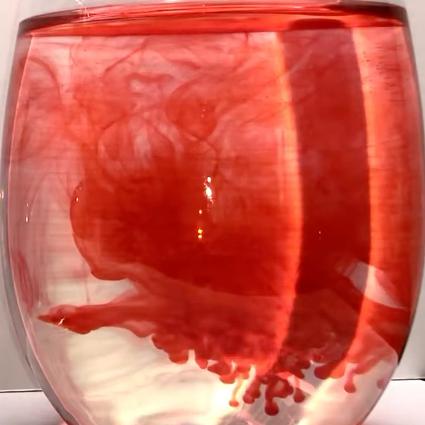}&
\includegraphics[width=0.1975\linewidth]{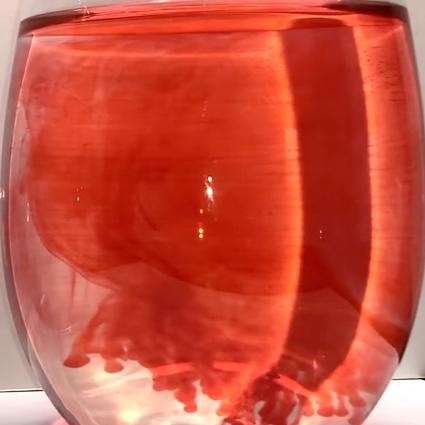}
\end{tabular}
}}
\vspace{-0.2in}
\caption{Illustrative examples of renormalization group (RG) flow. The aggregated pattern exhibits high multi-scale structural self-dissimilarity (MS$^3$D), denoted by $\mathcal{D}_\Gamma$.}
\label{fig:teaser}
\vspace{-0.2in}
\end{figure}

In this paper, we address the problem of GAN deterioration from a novel perspective and introduce a model regularization method. Unlike most existing regularization techniques~\cite{cvpr/TsengJL0Y21,aaai/Cui0LZZL22,cvpr/CuiYZLLX23} that focus on improving the discriminator's generalization across different data distributions to avoid overfitting, our approach explores the intrinsic properties of neural networks to uncover potential clues. Specifically, we observe that in settings with limited data, the gradients provided by the discriminator, i.e., $\nabla_{x}f(x;\phi)$, gradually exhibit an aggregation pattern. This pattern indicates that the discriminator concentrates its attention on a small portion of the input image, rather than capturing comprehensive details. This phenomenon, which we term the \emph{perceptual narrowing phenomenon}, varies in degree across different GANs and tasks and accompanies overfitting.

To understand this phenomenon, here, we draw inspiration from the renormalization group (RG) concept in physics~\cite{ppf/kadanoff1966,prb/wilson1971}, which progressively separates coarse-grained statistics from fine-grained statistics through local transformations at different scales. We apply this idea to analyze the gradient field $\nabla_{x}f(x;\phi)$ at different scales. Unlike scatter patterns, the aggregated pattern shows significant divergence from its coarse-grained versions, as depicted in Fig.~\ref{fig:teaser}. This \textit{self-dissimilarity} (SD)\footnote{A concept in complexity theory and the first formalization of SD appeared in~\cite{utics/wolpert2018}.} can be seen as the system's unique ``signature''~\cite{utics/wolpert2018,complexity/WolpertM07}, revealing how information processing changes across different scales in the system~\cite{ds/Jacobs1992,complexity/WolpertM07}. A high SD indicates that the system is efficient, encoding substantial processing into its dynamics, but also sensitive or unstable, where a small change can cause large dynamic shifts. Conversely, a low SD is associated with robustness, where a system can maintain functionality despite disturbances, being less efficient but more reliable due to built-in redundancies. The latter is more desirable for GAN training.

Based on this analysis, we introduce a \textbf{m}ulti-\textbf{s}cale \textbf{s}tructural \textbf{s}elf-\textbf{d}issimilarity (MS$^3$D) regularization based on RG flow. By repeatedly applying RG transformations, we generate a series of coarse-grained versions at different scales. Along the RG flow, we compute SD at each scale and combine them to obtain MS$^3$D. This regularization enforces the gradient field to maintain a similar pattern or structure across scales, promoting a more redundant and robust feedback system for the generator. These properties help avoid overfitting and collapse in GAN training with limited data. Crucially, the proposed MS$^3$D computation method is differentiable and can be readily implemented using popular deep learning frameworks like PyTorch~\cite{nips/PaszkeGMLBCKLGA19} and TensorFlow~\cite{corr/AbadiABBCCCDDDG16}. We verify the effectiveness of our method on various GANs and datasets. The results demonstrate that it improves GAN performance under limited data conditions in terms of generalization and stability. Notably, our method is orthogonal to data augmentation methods and other model constraint techniques, and can be integrated with them to further enhance GAN performance with very small datasets. Our work provides a new perspective on GAN training with limited data and offers a novel regularization method to improve GAN performance in this challenging scenario.

\section{Related Work}
\textbf{Generative adversarial networks.}
Generative adversarial networks (GANs)~\cite{nips/GoodfellowPMXWOCB14} have made significant advancements in generating high-quality and diverse images. This progress is attributed to the development of more robust objective functions~\cite{icml/ArjovskyCB17,nips/GulrajaniAADC17,iclr/ZhaoML17,iccv/MaoLXLWS17,icml/SongE20}, advanced architectures~\cite{iclr/MiyatoKKY18,iclr/MiyatoK18,icml/ZhangGMO19}, and effective training strategies~\cite{nips/DentonCSF15,iccv/ZhangXL17,iclr/KarrasALL18,cvpr/Liu0BZ020}. Notable examples include BigGAN~\cite{iclr/BrockDS19} and StyleGAN~\cite{cvpr/KarrasLA19,cvpr/KarrasLAHLA20}, which are capable of generating high-resolution images with rich details and diverse styles. This paper focuses on the challenges and solutions for training GANs with limited data.

\begin{figure*}[!htbp]
\centering
\subfigure[Discriminator outputs, FIDs]{
\centering
\includegraphics[width=0.246\linewidth]{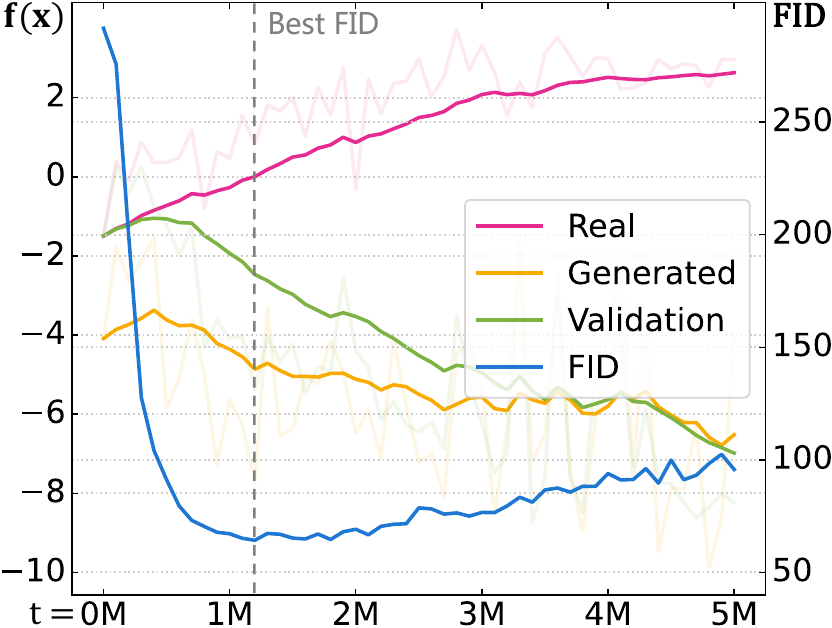}
}
\hspace{0.08in}
\vspace{-0.08in}
\subfigure[The aggregation degree of $\nabla_{x}f(x;\phi)$]{
\centering
\includegraphics[width=0.23\linewidth]{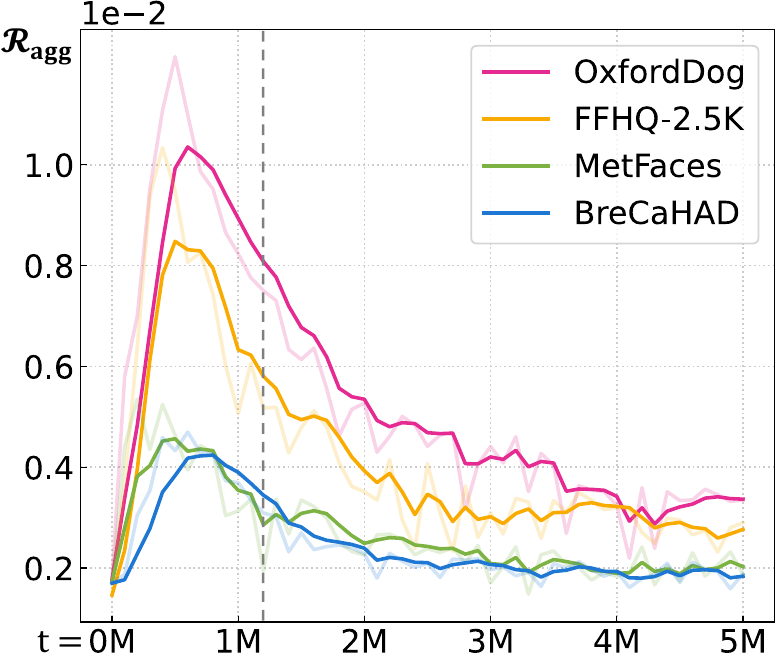}
\hspace{-0.04in}
\includegraphics[width=0.23\linewidth]{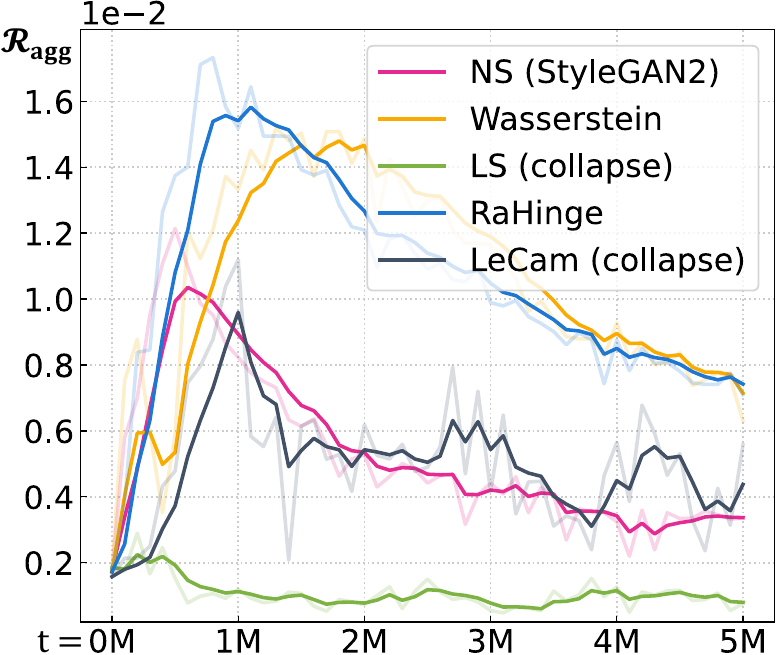}
\hspace{-0.04in}
\includegraphics[width=0.23\linewidth]{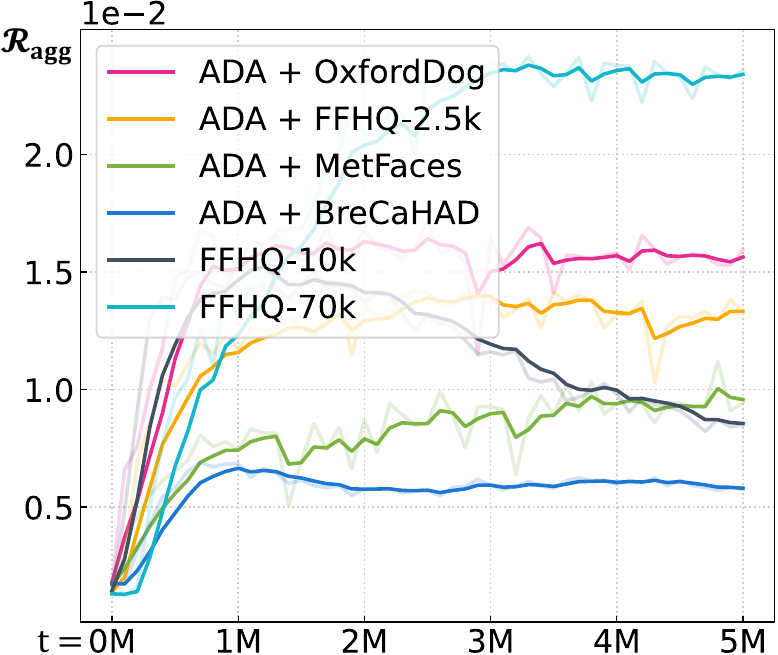}
}
\vspace{-0.08in}
\caption{(a) Evolution of discriminator outputs and FID values during training, illustrating the discriminator's overfitting and the resulting degradation of generated samples. (b) Consistent aggregation tendency of the gradient $\nabla_{x}f(x;\phi)$ across various GANs and datasets under limited data conditions.}

\label{fig:perceptual_narrowing}
\vspace{-0.1in}
\end{figure*}

\begin{figure}[t]
\setlength{\tabcolsep}{1pt}
\centering
{\small
\begin{tabular}{c c c c}
\vspace{-1pt}
\raisebox{0.095\linewidth}{\rotatebox[origin=t]{90}{1.2M}}&
\includegraphics[width=0.25\linewidth]{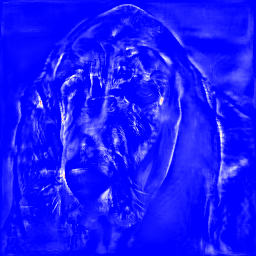}&
\includegraphics[width=0.25\linewidth]{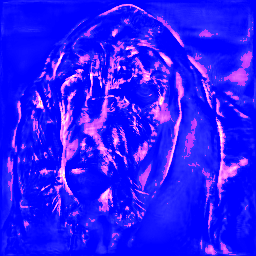}&
\includegraphics[width=0.25\linewidth]{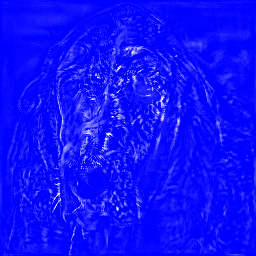}
\tabularnewline
\raisebox{0.095\linewidth}{\rotatebox[origin=t]{90}{5M}}&
\includegraphics[width=0.25\linewidth]{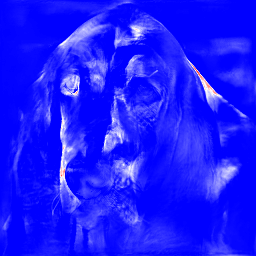}&
\includegraphics[width=0.25\linewidth]{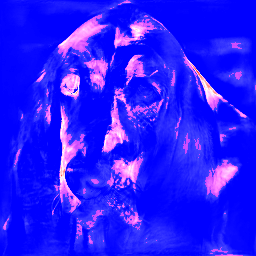}&
\includegraphics[width=0.25\linewidth]{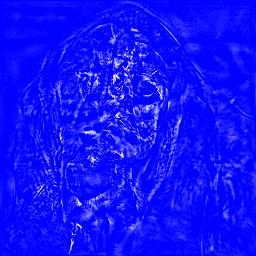}
\tabularnewline
&\multicolumn{2}{c}{(a) StyleGAN2} & (b) + ADA
\end{tabular}
}
\vspace{-0.08in}
\caption{Visualization of the gradient $\nabla_{x}f(x;\phi)$ at different training steps.}
\vspace{-0.2in}
\label{fig:connec}
\end{figure}

\textbf{Training GANs under limited data setting.}
Training GANs with limited data presents significant challenges, as the discriminator may overfit, leading to degraded generated samples~\cite{cvpr/WebsterRSJ19,iclr/GulrajaniRM19}. One common approach to mitigate this issue is data augmentation~\cite{nips/KarrasAHLLA20,corr/abs-2006-05338,iclr/ZhangZOL20,nips/ZhaoLLZ020,aaai/Zhao0LZOZ21,corr/abs-2006-02595}, which enriches the data distribution by applying transformations to the original samples.
However, data augmentation is not straightforward for GANs, as it can alter the target distribution or introduce artifacts (augmentation leaking). Recent methods have been designed to address these challenges, such as differentiable augmentation~\cite{nips/ZhaoLLZ020}, adaptive augmentation~\cite{nips/KarrasAHLLA20}, and generative augmentation~\cite{corr/abs-2006-02595}.

Another approach is model regularization, which prevents the discriminator from overfitting by imposing constraints or penalties on its parameters or outputs. While model regularization is commonly used to stabilize GAN training and prevent mode collapse, it is particularly effective in data-limited settings where overfitting is more severe. Techniques include adding noise to the discriminator's inputs or outputs~\cite{iclr/ArjovskyB17,iclr/SonderbyCTSH17,cvpr/JenniF19}, applying gradient penalties~\cite{nips/GulrajaniAADC17,icml/MeschederGN18}, using spectral normalization~\cite{iclr/MiyatoKKY18,iclr/MiyatoK18}, and adding consistency loss~\cite{iclr/ZhangZOL20}. Recent innovations, such as the LC regularization term~\cite{cvpr/TsengJL0Y21}, which modulates the discriminator's evaluations using two exponential moving average variables and connects to LeCam divergence~\cite{ssis/Cam1986}, have shown significant benefits. DigGAN~\cite{nips/Fang0S22} addresses gradient discrepancies between real and generated images, improving GAN performance. Additionally, leveraging external knowledge by using pre-trained models as additional discriminators~\cite{cvpr/Kumari0SZ22}, aligning discriminator features through knowledge distillation~\cite{cvpr/CuiYZLLX23}, or pre-training the highest resolution layer on larger datasets~\cite{corr/Sangwoo2020} has also proven effective.

Our work introduces a novel model regularization technique, offering deeper insights into GANs' internal dynamics through the lens of the renormalization group.

\textbf{Renormalization group and its applications.}
The renormalization group (RG) is a powerful tool in modern physics that elucidates how physical systems exhibit distinct behaviors at different observational scales~\cite{ppf/kadanoff1966,prb/wilson1971}. RG is extensively used in various physics fields, such as condensed matter theory~\cite{prb/wilson1971,romp/fisher1998,romp/stanley1999}, quantum field theory~\cite{npb/machacek1985}, and complexity theory~\cite{naos/bagrov2020}. Moreover, the concept of RG has been linked to information theory~\cite{np/koch2018,prl/gordon2021} and principal component analysis~\cite{josp/bradde2017}, making them more accessible to machine learning researchers. RG has also been employed to study neural networks' properties and dynamics. For example, ~\citet{corr/MehtaS14} demonstrated that RG can map a restricted Boltzmann machine to a hierarchical model, and ~\citet{josp/lin2017} showed that RG can explain the good generalization performance of deep neural networks. In this paper, we introduce the first application of RG to GAN training, proposing a novel regularization technique inspired by RG principles.

\begin{figure}[t]
\centering
\subfigure[]{
\centering
\includegraphics[width=0.48\linewidth]{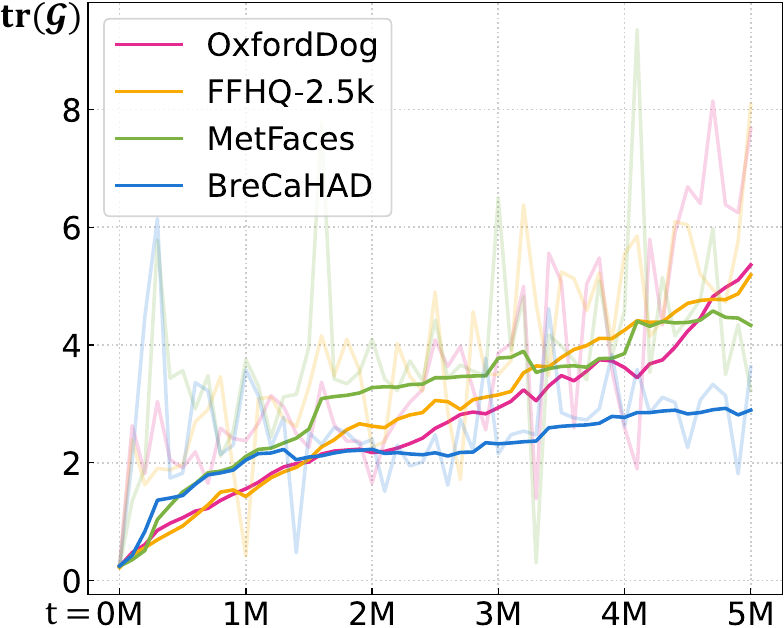}
}
\hspace{-0.08in}
\subfigure[]{
\centering
\includegraphics[width=0.48\linewidth]{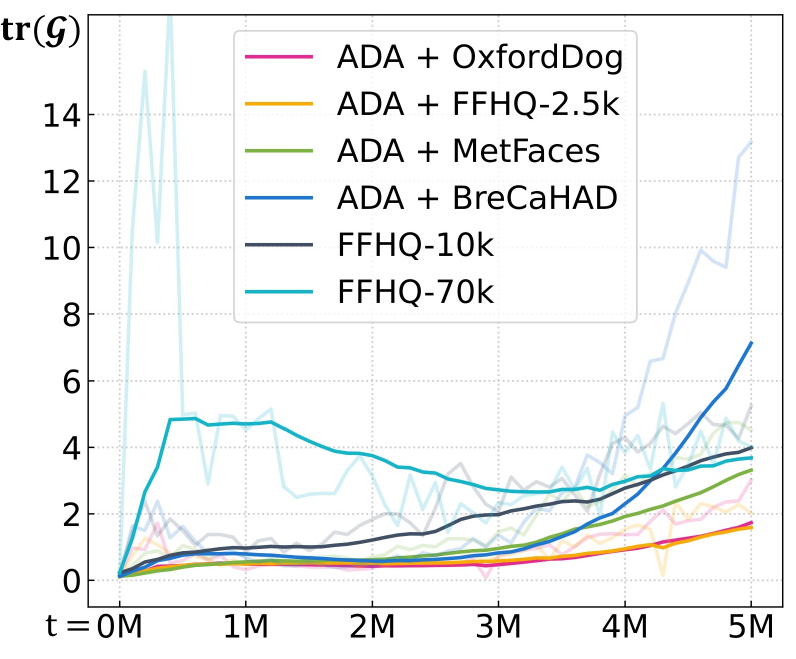}
}
\vspace{-0.2in}
\caption{(a) Under limited data settings, Fisher information increases during training, which indicates a decline in system stability. (b) When data augmentation is applied or the data volume is increased, Fisher information remains low, suggesting enhanced system stability.}
\label{fig:fisher}
\vspace{-0.1in}
\end{figure}

\section{Methodology}
\label{sec:methodology}
\subsection{Generative Adversarial Networks}
Generative adversarial networks (GANs)~\cite{nips/GoodfellowPMXWOCB14} are a class of generative models designed to synthesize realistic data samples from a latent noise vector \(z\). GANs consist of two neural networks: a generator \(g(\cdot;\theta)\) that transforms the noise vector into a data sample, and a discriminator \(f(\cdot;\phi)\) that distinguishes between real and generated samples. These networks are trained adversarially; the generator aims to produce samples that mimic the real data distribution, while the discriminator aims to accurately classify samples as real or fake. The training objective of GANs is formulated as a minimax game:
\begin{equation}
\min_\theta \max_\phi \mathbb{E}_{x}[\log f(x;\phi)] + \mathbb{E}_{z}[\log (1 - f(g(z;\theta);\phi))].
\end{equation}
The optimal solution is a Nash equilibrium where the generator produces samples indistinguishable from real data, and the discriminator assigns a probability of 0.5 to all samples.

\subsection{Perceptual Narrowing Phenomenon}
\label{sec:perceptual_narrowing}
Discriminator overfitting is a critical issue in GAN training with limited data, leading to a degradation in the quality of generated images~\cite{nips/ZhaoLLZ020,nips/KarrasAHLLA20,nips/JiangDWL21,cvpr/TsengJL0Y21}. This phenomenon is demonstrated in our experiments with StyleGAN2~\cite{cvpr/KarrasLAHLA20} on OxfordDog dataset~\cite{cvpr/ParkhiVZJ12}, as shown in Fig.~\ref{fig:perceptual_narrowing}(a). The discriminator becomes increasingly confident about the real images from the training set while becoming less confident about real images from the validation set, leading to a deterioration in Fr\'{e}chet Inception Distance (FID)~\cite{nips/HeuselRUNH17} over time.

Additionally, we observe that the gradient field \(\nabla_{x}f(x;\phi)\) of the discriminator with respect to the input becomes more aggregated over time, as shown in Fig.~\ref{fig:connec}(a). Intuitively, this aggregation may provide fragmented guidance to the generator. We refer to this as the \emph{perceptual narrowing phenomenon}. To quantify this, we devise a metric that counts the connected regions $N_\text{agg}$ by assigning 1 to gradient values above a threshold and 0 to the rest, followed by a connected component analysis. The ratio of connected regions to the total number of pixels, i.e., $\mathcal{R}_\text{agg}=\frac{N_\text{agg}}{H\times W}$, indicates the degree of gradient aggregation. Visual representations of connected regions are shown in Fig.~\ref{fig:connec}(a, right).

Extensive experiments on four small-scale datasets using various divergence measures and GAN architectures (detailed in Section~\ref{sec:experiments}) consistently showed that the number of connected regions initially increases but then decreases as training progresses, indicating increasing gradient aggregation (Fig.~\ref{fig:perceptual_narrowing}b, left). In contrast, with data augmentation or increased data volume to mitigate overfitting, the number of connected regions remains stable (Fig.~\ref{fig:perceptual_narrowing}b, middle and right). These findings suggest a link between overfitting and gradient aggregation in limited data GAN training.

Our analysis introduces a fresh perspective through the renormalization group (RG)~\cite{ppf/kadanoff1966,prb/wilson1971}, a technique devised to gradually extract the coarser statistical features of a physical system through local transformation. We observe that, in contrast to the dispersed pattern, the aggregated pattern demonstrates a significant divergence from its coarse-grained counterpart, as illustrated in Fig.~\ref{fig:teaser}. This self-dissimilarity (SD) reveals that the system processes information distinctively across different scales~\cite{ds/Jacobs1992,complexity/WolpertM07}, implying that the system is capable of encoding extensive information processing into its dynamics. This suggests that the system is both efficient and possesses a high degree of ``plasticity'' (learnability) from a neurological perspective~\cite{aron/Hensch2004}. Nonetheless, it also indicates that the system is sensitive and susceptible to minor disturbances that could lead to notable alterations in its dynamics~\cite{iclr/AchilleRS19}. These properties hint at the system's tendency toward overfitting and instability, aligning with our empirical findings. Conversely, a minimal SD points to the system's ability to generalize and maintain stability, which is favorable for GAN training.

To verify this point, we employ the Fisher information to examine the properties of deep neural networks during the training process. Although the connection form of weights in neural networks is fixed during training, not all weight connections contribute equally to the final output. We can view $\Psi(x;\phi) \triangleq \nabla_{x}f(x;\phi)$ as a mapping function parameterized by \(\phi\), encoding the posterior probability $p(y|x;\phi)$. The input is the generated or real image, and the output is the gradient of the discriminator with respect to the input, $y = \nabla_{x}f(x;\phi)$. By perturbing the weights and measuring the change in the output distribution using Kullback-Leibler (KL) divergence, we estimate the dependency of the final output on the weights. The second-order Taylor expansion of the KL divergence is:
\begin{equation}
\begin{aligned}
&\mathcal{D}_{KL}\left[p(y|x;\phi)||p(y|x;\phi+d\phi)\right] \\
=&\int_{\mathcal{X}} p(y|x;\phi)\log\left[\frac{p(y|x; \phi+d\phi)}{p(y|x;\phi)}\right]dx =\frac{1}{2} d\phi\cdot \mathcal{G} d\phi.
\end{aligned}
\end{equation}

\begin{lemma}~\cite{amari2016information}
\label{lem:metric}
Any standard f-divergence gives the same Riemannian metric $\mathcal{G}$, which is the Fisher information matrix (FIM)
    \begin{equation}
    \mathcal{G}=\mathbb{E}_{x, y}\left[\nabla_{\phi}\log p(y|x;\phi)\nabla_{\phi}\log p(y|x;\phi)^{\top}\right].
    \end{equation}
\end{lemma}
The FIM serves as a local measure to assess the effect of weight perturbations on outputs, reflecting the stability of the system \(\Psi(x;\phi)\). It also represents the strength of effective connectivity in neural networks and indicates their learnability~\cite{corr/KirkpatrickPRVD16}.

Following~\citet{iclr/AchilleRS19}, we use the diagonal FIM to reduce the computational cost:
\begin{equation}
tr(\mathcal{G})=\mathbb{E}_{x, y}\left[\Vert\nabla_{\phi}\log p(y|x;\phi)\Vert ^2\right].
\end{equation}

As illustrated in Fig.~\ref{fig:fisher}(a), Fisher information remains high and increases during the later stages of training, implying decreasing system stability and increasing learnability, hence a higher risk of overfitting and collapse. When the aggregation phenomenon is mitigated through augmentation or increased data volume, Fisher information declines during later training stages, as shown in Fig.~\ref{fig:fisher}(b). This aligns with our SD analysis and the CR-GAN approach~\cite{iclr/ZhangZOL20}, which uses semantically-preserving augmentation to enhance discriminator robustness by penalizing sensitivity to augmented images.

\subsection{Multi-scale Structural Self-dissimilarity}
Building upon the previous analysis, we introduce a novel regularization method, termed \textbf{m}ulti-\textbf{s}cale \textbf{s}tructural \textbf{s}elf-\textbf{d}issimilarity (MS$^3$D), designed to enhance the performance of GANs when trained with limited data. This method addresses the perceptual narrowing phenomenon by ensuring that the gradient field $\nabla_x(f(x;\phi))$ of the discriminator $f(x;\phi)$ maintains a consistent pattern across different scales, promoting a more redundant and robust feedback mechanism for the generator.

\textbf{Formulation.}
We quantitatively define MS$^3$D using the renormalization group (RG) flow concept. Consider the gradient $\nabla_x(f(x;\phi))$ of the discriminator with respect to the input $x$, denoted as $\Psi(x;\phi) \triangleq \nabla_x(f(x;\phi))$, where $\Psi(x;\phi)$ is parameterized by $\phi$ and belongs to the space $\mathcal{F}$ of real-valued functions defined on the Euclidean space $\mathcal{X}$.
An RG transformation $\Gamma$ can be naturally defined for $\Psi(x;\phi)$. In our discrete pixel setting $\mathcal{X}$, $\Gamma$ can be implemented using transformations like the Kadanoff block-spin transformation~\cite{ppf/kadanoff1966} or Gaussian filtering. Applying the RG transformation repeatedly results in a coarse-grained function $\Psi(x;\phi)$, generating the evolution of RG, that is, the RG flow.

In the following text, for clarity, let's denote the original system as $\Gamma_{0\to 0}(\Psi(x;\phi))\triangleq\Psi(x;\phi)$ and the coarse-grained system at scale $s$ as $\Gamma_{0\to s}(\Psi(x;\phi))$. The RG transformation chain is constructed as $\Gamma_{0\to s} = \Gamma_{s-ds\to s} \circ \cdots \circ \Gamma_{0\to ds} \circ \Gamma_{0\to 0}$, where $ds$ represents the scale step, and $\circ$ denotes the composition operator. In other words, $\Gamma_{0\to ds} \circ \Gamma_{0\to 0}(\cdot) = \Gamma_{0\to ds}(\Gamma_{0\to 0}(\cdot))$. For simplicity, we'll use $\Gamma_{0}\triangleq\Gamma_{0\to 0}$, $\Gamma_{s}\triangleq\Gamma_{0\to s}$, and so on. The coarser-grained version of $\Gamma_{s}(\Psi(x;\phi))$ can be denoted as $\Gamma_{s\to s+ds}\circ\Gamma_{s}(\Psi(x;\phi))$ or simply as $\Gamma_{s+ds}(\Psi(x;\phi))$. If there is a difference between the coarse-grained system $\Gamma_{s}(\Psi(x;\phi))$ and its coarser version $\Gamma_{s\to s+ds}(\Gamma_{s}(\Psi(x;\phi)))$, we can say that the scale $s$ contributes some information processing to the system. To measure this dissimilarity, we introduce self-dissimilarity (SD) as:
\begin{equation}
\label{eq:sd}
\begin{aligned}
&\mathcal{D}_{\Gamma_{s\to s+ds}} \\
=&\left|\langle\Gamma_{s}|\Gamma_{s+ds}\rangle-\frac{1}{2}(\langle\Gamma_{s}|\Gamma_{s}\rangle+\langle\Gamma_{s+ds}|\Gamma_{s+ds}\rangle) \right|  \\
=&\frac{1}{2}\int_{\mathcal{X}}\left (\Gamma_{s+ds}\left (\Psi(x;\phi)\right )-\Gamma_{s}(\Psi(x;\phi))\right )^2dx,
\end{aligned}
\end{equation}
where $\langle\Gamma_0|\Gamma_1\rangle=\int_{\mathcal{X}}\Gamma_0\Gamma_1dx$ represents the overlap between two scales.
In the context of Kadanoff decimation, SD is calculated by $\frac{1}{2}\left|(\langle\Gamma_{s}|\Gamma_{s}\rangle - \langle\Gamma_{s+ds}|\Gamma_{s+ds}\rangle)\right|$ (See the proof in Appendix A).
Along the RG flow, the multi-scale structural self-dissimilarity (MS$^3$D) is defined as the integration of SD across scales:
\begin{equation}
\mathcal{D}_{\Gamma_{0\to s}}=\sum_{i=0}^{s/ds}\mathcal{D}_{\Gamma_{i\to i+ds}}.
\end{equation}

\begin{figure}[t]
\centering
\includegraphics[width=\linewidth]{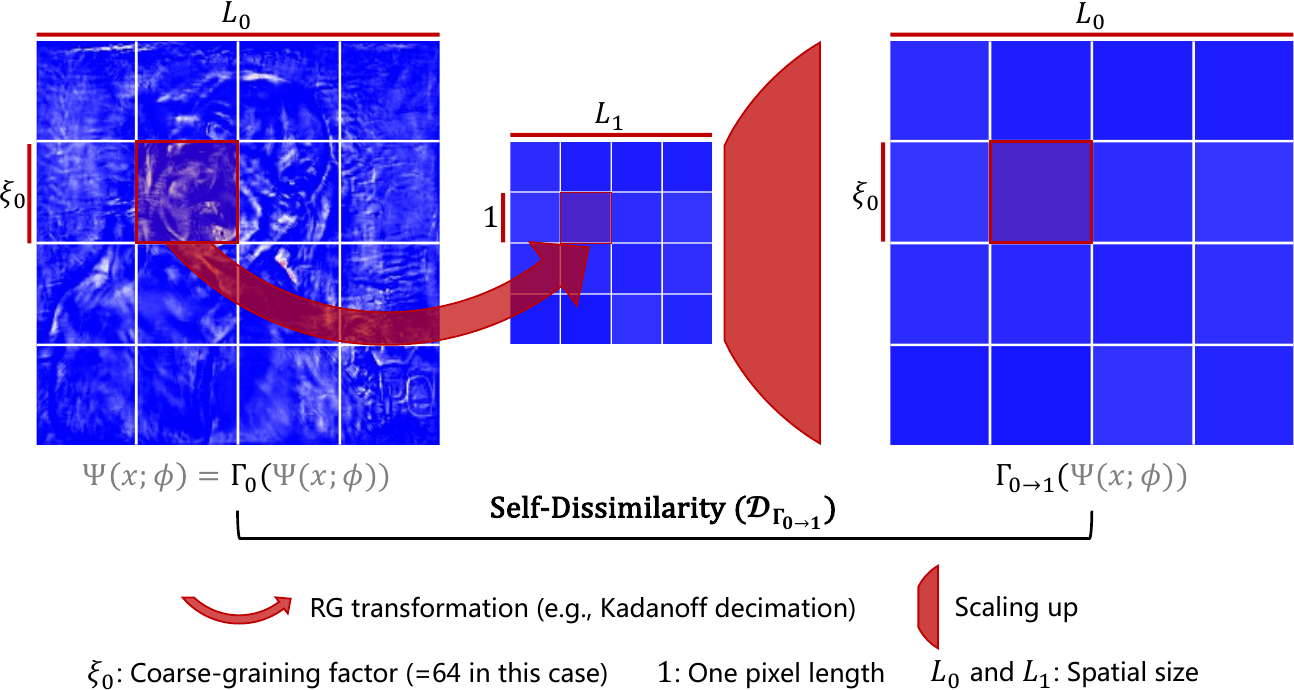}
\vspace{-0.2in}
\caption{A diagram illustrating the RG transformation process and MS$^3$D computation. It involves iteratively downsampling $\nabla_{x}f(x;\phi)$ and computing differences before and after downsampling.
}
\label{fig:rg_trans}
\vspace{-0.1in}
\end{figure}

\textbf{Implementation.}
We calculate the gradient of the discriminator's logits, denoted as $f(x;\phi)$, with respect to the input image $x$ using the equation $\Psi(x;\phi) = \frac{\partial f(x;\phi)}{\partial x}$. Here, $x$ is an image in the real-valued space $\mathbb{R}^{h \times w \times c}$, where $h$, $w$, and $c$ represent the height, width, and number of channels, respectively. The gradient field $\Psi(x;\phi): \mathbb{R}^{h \times w \times c} \to \mathbb{R}^{h \times w \times c}$ is then transformed into a square matrix $\Psi(x;\phi): \mathbb{R}^{L \times L} \to \mathbb{R}^{L \times L}$, where $L = \sqrt{h \times w \times c}$. To ensure that the matrix is square, we apply zero-padding to $\Psi(x;\phi)$, resulting in a square matrix $\tilde{\Psi}(x;\phi)$ of size $L \times L$.

For the real function $\tilde{\Psi}(\cdot;\phi)$ and the input image $x_0$, we normalize the output matrix $\tilde{\Psi}(x_0;\phi)$ to the range $[0, 1]$:
\begin{equation}
\tilde{\Psi}(x;\phi) \leftarrow \left| \frac{\tilde{\Psi}(x_0;\phi)}{\max(|\tilde{\Psi}(x_0;\phi)|)} \right|.
\end{equation}
We then use the Kadanoff block-spin transformation to coarse-grain the system, as illustrated in Fig.~\ref{fig:rg_trans}. Although the Kadanoff decimation is the simplest renormalization group (RG) transformation, it yields meaningful and robust outcomes, as verified in our ablation study (detailed in Section~\ref{sec:experiments}).
During the transformation $\Gamma_{s\to s+ds}$, the matrix $\Gamma_s(\tilde{\Psi}(x_0;\phi))$, with spatial dimensions $L_s\times L_s$, is tiled by blocks of size $\zeta_s\times \zeta_s$, where $\zeta$ is the coarse-graining factor. In this paper, we set $\zeta_0=\zeta_1=\cdots=\zeta_s=2$.
The coarse-grained matrix $\Gamma_{s+ds}(\tilde{\Psi}(x_0;\phi))$ is obtained by replacing each block with its average value. This matrix retains the original spatial dimensions $L_{s+ds}=L_s$, but the number of elements is reduced by a factor of $(\zeta_s)^2$. Mathematically, this process is represented as follows:
\begin{equation}
\label{eq:coarse_grain}
\begin{split}
&\left\{\Gamma_{s+ds}(\tilde{\Psi}(x_0;\phi))\right\}(i,j) \\
=&\frac{1}{(\zeta_{s})^2}\sum_{m=0}^{\zeta_{s}-1}\sum_{n=0}^{\zeta_{s}-1}\left\{\Gamma_{s}(\tilde{\Psi}(x_0;\phi))\right\}\left (\lfloor\frac{i}{\zeta_{s}}\rfloor\cdot\zeta_{s}+m, \right. \\
&\left.\lfloor\frac{j}{\zeta_{s}}\rfloor\cdot\zeta_{s}+n\right ),
\end{split}
\end{equation}
where $\lfloor\cdot\rfloor$ is the floor function, and $\left\{\Gamma(\cdot)\right\}(i,j)$ indicates the element at the $i$-th row and $j$-th column of $\Gamma(\cdot)$.

Subsequently, the matrix $\Gamma_{s+ds}(\tilde{\Psi}(x_0;\phi))$ is tiled by blocks of size $\zeta^{s+ds}\times \zeta^{s+ds}$, and its coarser-grained version $\Gamma_{s+2ds}(\tilde{\Psi}(x_0;\phi))$ is obtained by replacing each block with its average value. This process is repeated until $(\zeta)^{t+1} > L$, i.e., the coarse-grained transformation cannot be applied anymore, where $t$ is the number of iterations and $1\leq t\leq\log_{\zeta}L$. For our purposes, we set $\zeta=2$.
To integrate the MS$^3$D regularization within widely-used computational frameworks like PyTorch~\cite{nips/PaszkeGMLBCKLGA19} and TensorFlow~\cite{corr/AbadiABBCCCDDDG16}, we utilize a convolution operation with an average filter of size $2\times 2$ and a stride of 2 to perform the Kadanoff decimation. The coarse-grained output is then scaled back to the original dimensions by a factor of 2. In Appendix B, we present the PyTorch-like pseudo-code of MS$^3$D calculation.
The dissimilarity between the two versions is computed by the overlap of the current version and the coarser-grained version (Eq.~(\ref{eq:sd})). We simply implement SD as the mean squared error (MSE). Hence, the MS$^3$D is the cumulative sum of SDs across all scales, formulated as:
\begin{equation}
    \mathcal{D}_{\Gamma_{0 \to t}} = \sum_{i=0}^{t} \|\Gamma_{i}(\Psi(x_0;\phi)) - \Gamma_{i \to i+1}(\Psi(x_0;\phi))\|^2_2.
\end{equation}

The loss function for the discriminator is:
\begin{equation}
\label{eq:loss_dis}
\begin{aligned}
\mathcal{L}_{\text{dis}}=&-\mathbb{E}_{x}\log f(x;\phi)-\mathbb{E}_{z}\log (1-f(g(z));\phi) \\
&+\lambda\mathbb{E}_{x}\mathcal{D}_{\Gamma_{0\to t}}(\nabla_x(f(x;\phi))),        
\end{aligned}
\end{equation}
where $\lambda$ is the weight of the MS$^3$D regularization term.

\begin{table}[t]
\caption{Performance comparison of various GAN loss functions (divergence measures). $\Uparrow$ denotes that higher values are preferable, while $\Downarrow$ indicates that lower values are better.}
\label{tab:comp_losses}
\begin{center}
\begin{small}
\begin{sc}
\resizebox*{0.4\textwidth}{!}{
\begin{tabular}{lrrrr}
\toprule
\multirow{2}{*}{\textbf{Method}} & \multicolumn{2}{c}{\textbf{OxfordDog}} & \multicolumn{2}{c}{\textbf{FFHQ-2.5K}} \\
\cmidrule(r){2-3} \cmidrule(r){4-5}
& FID $\Downarrow$ & IS $\Uparrow$ & FID $\Downarrow$ & IS $\Uparrow$ \\
\midrule
StyleGAN2 (NS) & 64.26 & 9.69 & 48.11 & 3.50  \\
+ Wasserstein & 82.18 & 9.94 & 38.64 & 4.11  \\
+ LS & 216.42 & 2.69 & 213.93 & 2.17  \\
+ RaHinge & 38.68 & 11.14 & 32.34 & 3.96  \\
\rowcolor{mygray} + MS$^3$D & 47.05 & 10.73 & 33.46 & 4.61  \\
\rowcolor{mygray} + MS$^3$D + RaHinge & \textbf{37.71} & \textbf{12.54} & \textbf{31.95} & \textbf{4.70}  \\
\bottomrule
\end{tabular}
}
\end{sc}
\end{small}
\end{center}
\vspace{-0.15in}
\end{table}

\section{Experiments}
\label{sec:experiments}

\subsection{Experimental Settings}
\textbf{Datasets.}
Our experiments utilize four diverse datasets: OxfordDog (from Oxford-IIIT pet dataset~\cite{cvpr/ParkhiVZJ12}, detailed in Appendix C), Flickr-Faces-HQ (FFHQ)~\cite{cvpr/KarrasLA19}, MetFaces~\cite{nips/KarrasAHLLA20}, and BreCaHAD~\cite{brn/Aksac2019}. Detailed descriptions of these datasets are provided in Appendix C.

\textbf{Metrics.}
We employ Inception Score (IS)~\cite{nips/SalimansGZCRCC16}, Fr\'{e}chet Inception Distance (FID)~\cite{nips/HeuselRUNH17}, and Kernel Inception Distance (KID)~\cite{iclr/BinkowskiSAG18} to evaluate our models. We utilize the official implementations of these metrics as provided by \citet{cvpr/KarrasLAHLA20}.

\begin{table}[t]
\caption{Comparison of techniques for mitigating discriminator overfitting under limited data.}
\label{tab:comp_reg}
\begin{center}
\begin{small}
\begin{sc}
\resizebox*{0.36\textwidth}{!}{
\begin{tabular}{lrrrr}
\toprule
\multirow{2}{*}{\textbf{Method}} & \multicolumn{2}{c}{\textbf{OxfordDog}} & \multicolumn{2}{c}{\textbf{FFHQ-2.5K}} \\
\cmidrule(r){2-3} \cmidrule(r){4-5}
& FID $\Downarrow$ & IS $\Uparrow$ & FID $\Downarrow$ & IS $\Uparrow$ \\
\midrule
StyleGAN2 & 64.26 & 9.69 & 48.11 & 3.50  \\
+ LeCam & 102.87 & 8.02 & 68.85 & 3.53  \\
+ DigGAN & 61.84 & 10.07 & 48.45 & 3.68  \\
+ KD-DLGAN & 54.06 & 9.73 & 49.31 & 3.67 \\
\rowcolor{mygray}+ MS$^3$D & 47.05 & 10.73 & 33.46 & 4.61  \\
+ CR + ADA & 29.81 & 11.76 & \textbf{19.98} & 4.53  \\
\rowcolor{mygray}+ MS$^3$D + ADA & \textbf{25.94} & \textbf{12.08} & 20.23 & \textbf{4.71}  \\
\bottomrule
\end{tabular}
}
\end{sc}
\end{small}
\end{center}
\vspace{-0.2in}
\end{table}

\subsection{Results: Various Limited Data Settings}
In Table~\ref{tab:comp_losses}, we compare our method with multiple GAN losses, including Wasserstein~\cite{icml/ArjovskyCB17}, Least Squares (LS)~\cite{iccv/MaoLXLWS17}, RaHinge~\cite{iclr/Jolicoeur-Martineau19}, and NS (non-saturated, vanilla StyleGAN2)~\cite{nips/GoodfellowPMXWOCB14}. We observe that the state-of-the-art divergence metrics, such as Wasserstein and LS, which are originally devised to stabilize GAN training, perform poorly in limited data settings. Surprisingly, RaHinge loss performs very well, unlike the poor results reported by~\citet{cvpr/TsengJL0Y21}, and it outperforms our default setting, i.e., StyleGAN2 + MS$^3$D (abbreviated as + MS$^3$D). A possible reason is that the relativistic discriminator reduces the probability of real data being classified as real, thus enhancing the discriminator's generalization ability. Nevertheless, our method still surpasses it and achieves the best performance on + MS$^3$D + RaHinge. This indicates that our method can effectively integrate with other methods.

\begin{table}[t]
\caption{Comparison of various GAN architectures and techniques.}
\label{tab:comp_arc}
\begin{center}
\begin{small}
\begin{sc}
\resizebox*{0.38\textwidth}{!}{
\begin{tabular}{llrrrr}
\toprule
\textbf{Method} & & \textbf{FID} $\Downarrow$ & \textbf{IS} $\Uparrow$  \\
\midrule
\multirow{5}{*}{DCGAN} & Vanilla & 107.33 & 3.23  \\
& + Auxiliary rotations & 103.43 & 2.85  \\
& + Dropout & 77.35 & 4.11  \\
& + Noise & 86.13 & \textbf{5.40}  \\
\rowcolor{mygray}& + MS$^3$D & \textbf{59.37} & 4.39  \\
\midrule
\multirow{5}{*}{SNGAN} & Vanilla & 57.08 & 3.42  \\
& + Auxiliary rotations & 54.85 & 3.63  \\
& + Dropout & 48.54 & 4.02  \\
& + Noise & 55.99 & 3.33  \\
\rowcolor{mygray}& + MS$^3$D & \textbf{47.87} & \textbf{4.18}  \\

\bottomrule
\end{tabular}
}
\end{sc}
\end{small}
\end{center}
\vspace{-0.2in}
\end{table}

In Table~\ref{tab:comp_reg}, we show the comparison with other regularization methods, namely: consistency regularization (CR)~\cite{iclr/ZhangZOL20}, exponential moving average regularization (LeCam)~\cite{cvpr/TsengJL0Y21}, discriminator gradient gap regularization (DigGAN)~\cite{nips/Fang0S22}, and knowledge distillation (KD-DLGAN)~\cite{cvpr/CuiYZLLX23}.
CR aims to constrain the discriminator's sensitivity to augmented data, relying on augmentation, while our method is augmentation-free. To provide a comprehensive comparison, we also combine adaptive discriminator augmentation (ADA)~\cite{nips/KarrasAHLLA20} with our method. As shown in Table~\ref{tab:comp_reg}, our method outperforms LeCam, DigGAN, and KD-DLGAN, which are specifically designed for limited data scenarios. Additionally, in combination with ADA (+ MS$^3$D + ADA), our method surpasses CR with ADA (+ CR + ADA).

Furthermore, we compare our method across different GAN architectures, including DCGAN~\cite{corr/RadfordMC15} and SNGAN~\cite{iclr/MiyatoKKY18}, and various improvement techniques such as auxiliary rotation prediction~\cite{cvpr/ChenZRLH19}, dropout~\cite{jmlr/SrivastavaHKSS14}, and noise injection~\cite{iclr/SonderbyCTSH17}. As demonstrated in Table~\ref{tab:comp_arc}, our method performs well not only with StyleGAN architectures but also with traditional convolution-based DCGAN and ResNet-based SNGAN, indicating that our method's effectiveness is not dependent on the underlying neural network structure, likely due to its primary impact on gradients.

\begin{figure}[t]
\centering
\subfigure[OxfordDog]{
\centering
\includegraphics[width=0.48\linewidth]{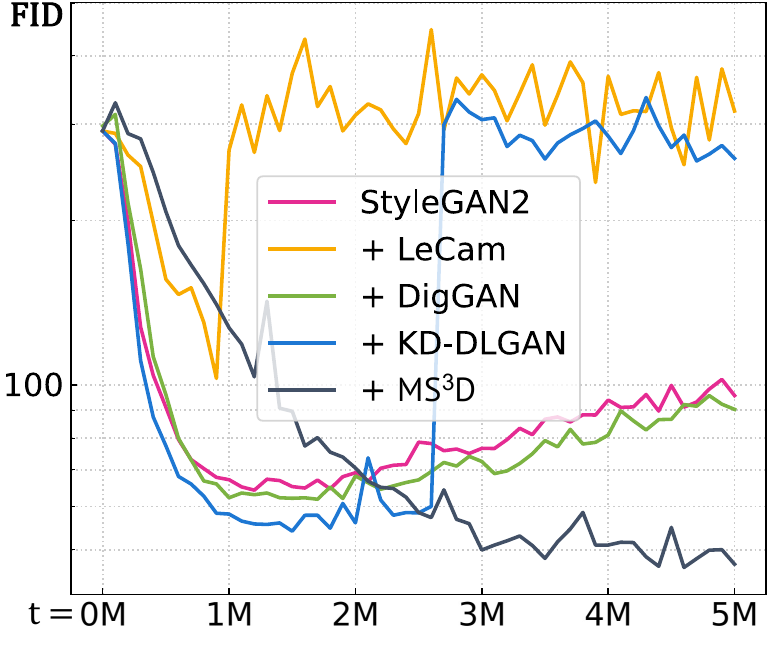}
}
\hspace{-0.08in}
\subfigure[FFHQ-2.5K]{
\centering
\includegraphics[width=0.48\linewidth]{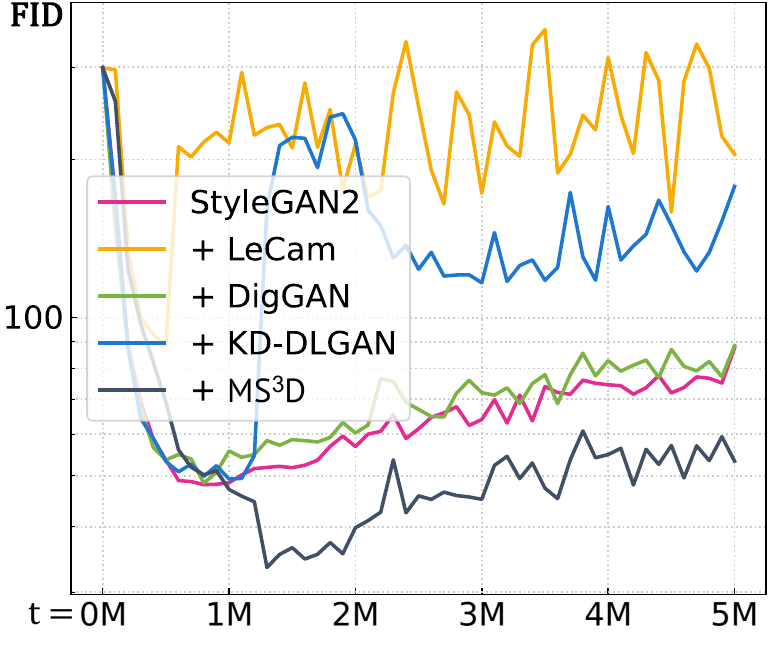}
}
\vspace{-0.15in}
\caption{Evolution of FID values on OxfordDog and FFHQ-2.5K datasets over various training steps.}
\label{fig:fid_over_step}
\vspace{-0.1in}
\end{figure}

To further demonstrate the effectiveness of our method in improving the quality of generated images under limited data conditions, we present FID values at various data scales in Table~\ref{tab:comp_scale}. Our method significantly reduces FID values, particularly at more limited data scales such as 1K, 2K and 4K.
In Fig.~\ref{fig:fid_over_step}, we illustrate the FID values over training steps for OxfordDog and FFHQ-2.5K. These results confirm that our method effectively mitigates the image quality deterioration typically observed with limited data. The generated images, as depicted in Fig.~\ref{fig:main_qualitative}, visibly demonstrate that our method yields more realistic images in comparison to baseline models.

\begin{table}[tbp]
\caption{Quantitative evaluation (FID $\Downarrow$) on varying data scales of the FFHQ dataset.}
\label{tab:comp_scale}
\begin{center}
\begin{small}
\begin{sc}
\resizebox*{\linewidth}{!}{
\begin{tabular}{lrrrrrrrr}
\toprule
\textbf{Method} & \textbf{1K} & \textbf{2K} & \textbf{4K} & \textbf{6K} & \textbf{8K} & \textbf{10K} & \textbf{70K} \\
\midrule
StyleGAN2 & 87.86 & 54.96 & 27.63 & 22.97 & 15.98 & 12.50 & 6.50  \\
+ LeCam & 99.90 & 88.51 & 69.55 & 33.46 & 16.06 & \textbf{10.48} & 6.53  \\
+ DigGAN & 81.42 & 56.33 & 31.26 & 20.92 & 15.77 & 12.61 & 6.63  \\
+ KD-DLGAN & 83.67 & 55.34 & 29.91 & \textbf{19.84} & 15.99 & 12.73 & 7.11  \\
+ RaHinge & 93.14 & 56.64 & 25.72 & 20.43 & \textbf{12.55} & 10.98 & \textbf{6.10}  \\
\rowcolor{mygray}+ MS$^3$D & \textbf{81.30} & \textbf{49.17} & \textbf{24.17} & 19.86 & 12.61 & 10.59 & 6.28  \\
\bottomrule
\end{tabular}
}
\end{sc}
\end{small}
\end{center}
\vspace{-0.2in}
\end{table}

\begin{table}[!tbp]
\caption{Quantitative comparison on small datasets, MetFaces and BreCaHAD.}
\label{tab:comp_tech_small}
\begin{center}
\begin{small}
\begin{sc}
\resizebox*{\linewidth}{!}{
\begin{tabular}{lrrrrrr}
\toprule
\multirow{2}{*}{\textbf{Method}} & \multicolumn{3}{c}{\textbf{MetFaces}} & \multicolumn{3}{c}{\textbf{BreCaHAD}} \\
\cmidrule(r){2-4} \cmidrule(r){5-7}
& FID $\Downarrow$ & IS $\Uparrow$ & KID $\Downarrow$ & FID $\Downarrow$ & IS $\Uparrow$ & KID $\Downarrow$ \\
\midrule
StyleGAN2 & 53.21 & 3.16 & 0.028 & 97.06 & \textbf{3.10} & 0.095  \\
+ LeCam & 56.67 & 2.43 & 0.102 & 83.74 & 2.51 & 0.046  \\
+ DigGAN & 53.97 & 3.09 & 0.031 & 105.45 & 3.05 & 0.095  \\
+ KD-DLGAN & 54.05 & 3.11 & 0.029 & 86.25 & 3.26 & 0.083  \\
\rowcolor{mygray}+ MS$^3$D & 37.71 & 3.85 & 0.017 & 67.12 & 2.62 & 0.038  \\
+ CR + ADA & 29.91 & 4.20 & 0.008 & 22.69 & 2.81 & 0.011  \\
\rowcolor{mygray}+ MS$^3$D + ADA & 23.09 & 4.15 & \textbf{0.004} & \textbf{20.22} & 2.81 & \textbf{0.006}  \\
\rowcolor{mygray}+ MS$^3$D + ADA + CR & \textbf{21.54} & \textbf{4.28} & \textbf{0.004} & 21.09 & 2.92 & 0.007  \\
\bottomrule
\end{tabular}
}
\end{sc}
\end{small}
\end{center}
\vspace{-0.2in}
\end{table}

\subsection{Results: Small Dataset}
We also conducted experiments on small datasets, MetFaces and BreCaHAD, which contain only 1,203 and 1,750 training images, respectively. Table~\ref{tab:comp_tech_small} compares our method with other techniques on these datasets, demonstrating that our method achieves the best performance. Following \citet{nips/KarrasAHLLA20}, we use KID for quantitative evaluation due to its stability on small datasets. The generated images in Fig.~\ref{fig:main_qualitative} show that our method produces more realistic images than baseline models.

\begin{table}[!tbp]
\caption{Quantitative results of transfer learning with Freeze-D technique applied to MetFaces dataset. The model is pretrained on FFHQ-70K.}
\label{tab:comp_transfer}
\begin{center}
\begin{small}
\begin{sc}
\resizebox*{0.32\textwidth}{!}{
\begin{tabular}{lrrr}
\toprule
\multirow{2}{*}{\textbf{Method}} & \multicolumn{3}{c}{\textbf{MetFaces}} \\
\cmidrule(r){2-4}
& FID $\Downarrow$ & IS $\Uparrow$ & KID $\Downarrow$ \\
\midrule
StyleGAN2 & 53.21 & 3.16 & 0.028  \\
+ Freeze-D & 47.74 & 3.32 & 0.020  \\
\rowcolor{mygray}+ MS$^3$D + Freeze-D & \textbf{36.95} & \textbf{3.56} & \textbf{0.013}  \\
\bottomrule
\end{tabular}
}
\end{sc}
\end{small}
\end{center}
\vspace{-0.2in}
\end{table}

\begin{figure*}[t]
\centering
\subfigure[]{
\centering
\includegraphics[width=0.22\linewidth]{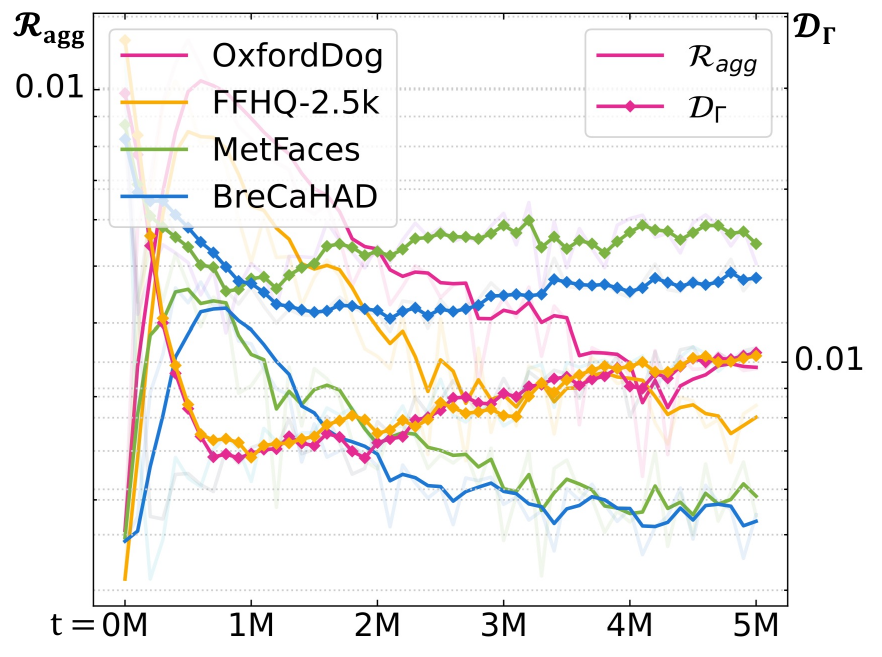}
}
\hspace{-0.08in}
\subfigure[]{
\centering
\includegraphics[width=0.2\linewidth]{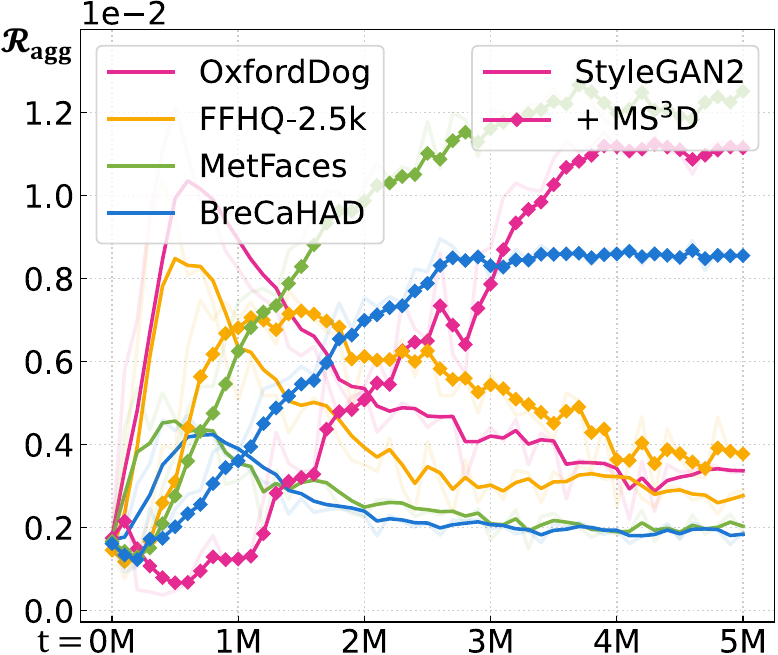}
}
\subfigure[]{
\includegraphics[width=0.082\linewidth]{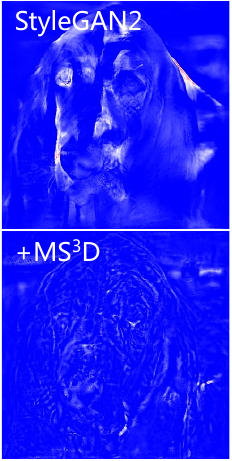}
}
\subfigure[]{
\includegraphics[width=0.206\linewidth]{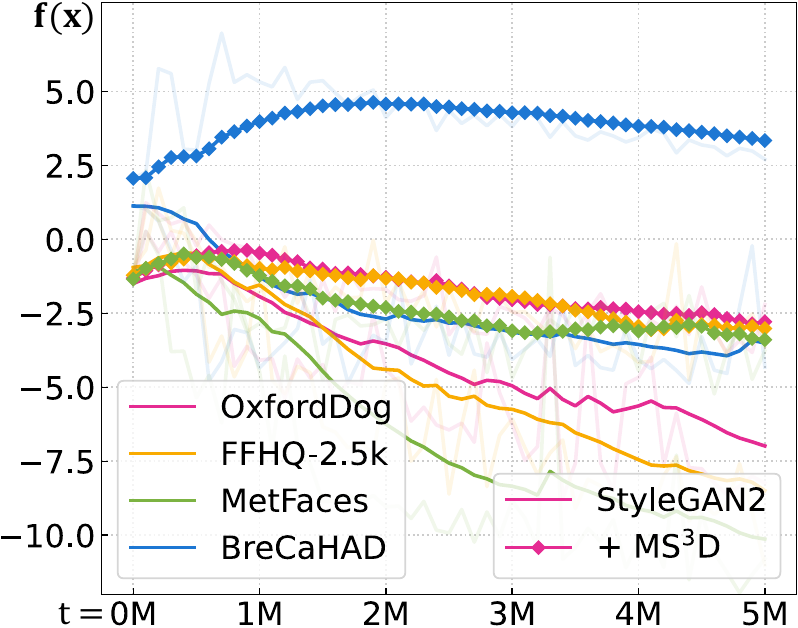}
}
\subfigure[]{
\includegraphics[width=0.203\linewidth]{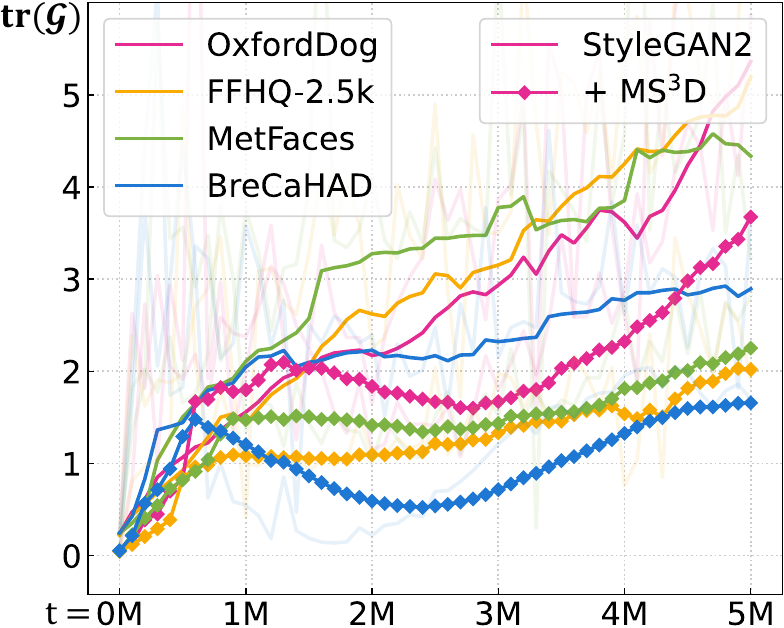}
}
\vspace{-0.15in}
\caption{(a) The relationship between perceptual narrowing and MS$^3$D. (b) Reduction in gradient aggregation post-MS$^3$D application. (c) Visualization of gradients post-MS$^3$D, showing a dispersed pattern similar to augmentation methods. (d) Mitigation of discriminator overfitting to validation data with MS$^3$D. (e) Decrease in Fisher information post-MS$^3$D, indicating enhanced stability in training dynamics.}
\label{fig:ratio_down}
\vspace{-0.1in}
\end{figure*}

\subsection{Results: Transfer Learning}
To cope with the difficulty of data collection, recent researchers have leveraged transfer learning in the image generation setting. \citet{cvpr/WangGBHK020} use fine-tuning to transfer the knowledge of models pre-trained on external large-scale datasets. Some works propose to fine-tune only part of the model, such as \citet{corr/Sangwoo2020} propose to fix the highest resolution layer of the discriminator (Freeze-D) for knowledge transfer, showing competitive results. In Table~\ref{tab:comp_transfer}, we implement this technique and show the results. We find that our method achieves consistent gains on all datasets, although the gains are very marginal.

\begin{figure}
\setlength{\tabcolsep}{1pt}
\centering
\subfigure[Loss landscapes]{
\includegraphics[width=0.33\linewidth]{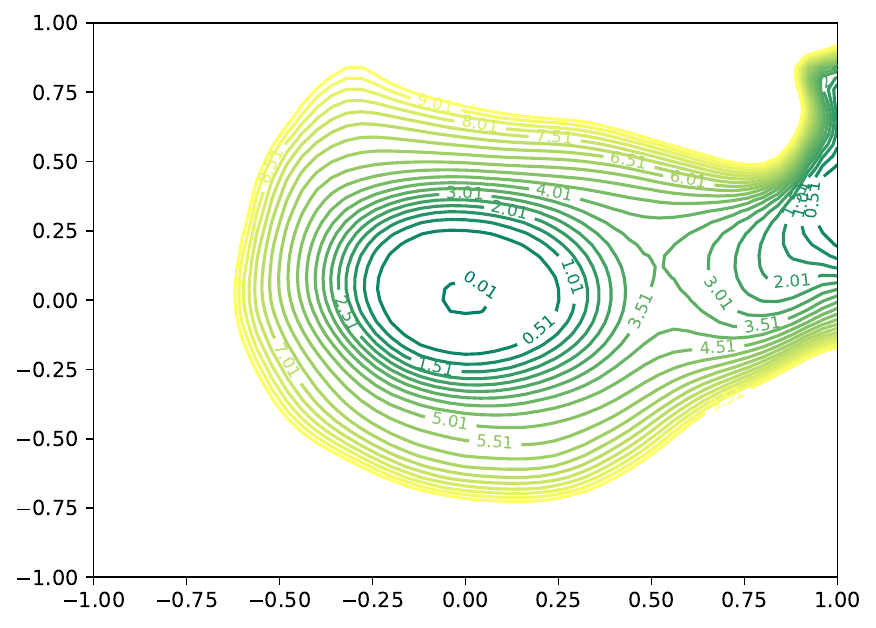}
\includegraphics[width=0.33\linewidth]{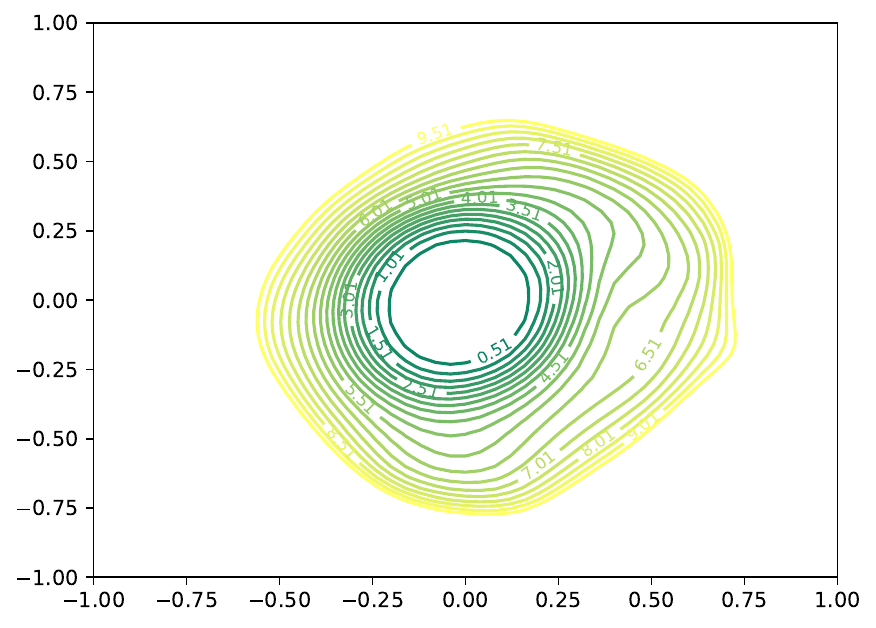}
}
\hspace{-0.05in}
\subfigure[Similarities]{
\includegraphics[width=0.28\linewidth]{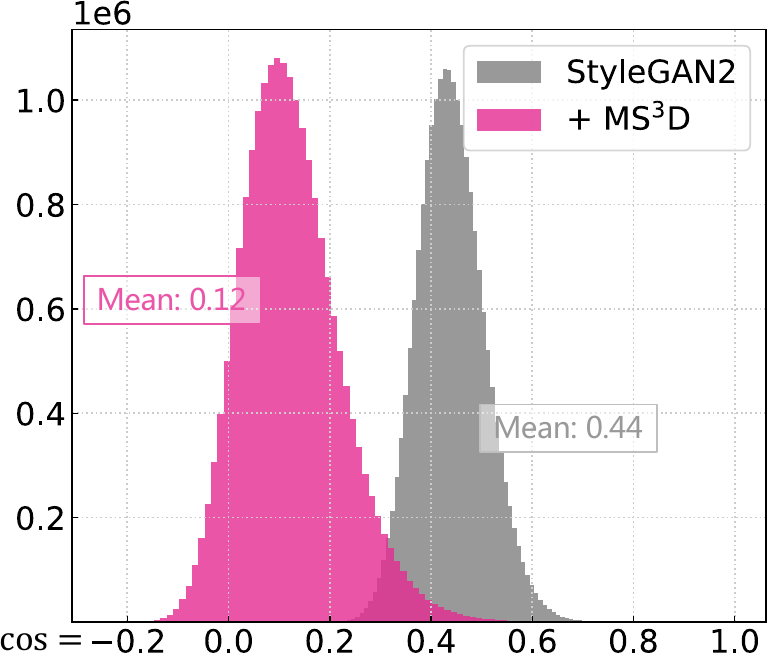}
}
\vspace{-0.15in}
\caption{(a) Loss landscapes of the discriminator without MS$^3$D (left) and with MS$^3$D (right). (b) Average cosine similarity of the discriminator's data embeddings.}
\label{fig:landscape_sim}
\vspace{-0.05in}
\end{figure}

\begin{table}[t]
\caption{Computational complexity of different methods.}
\label{tab:comp_complexity}
\begin{center}
\begin{small}
\begin{sc}
\resizebox*{\linewidth}{!}{
\begin{tabular}{lccc}
\toprule
\textbf{Method} & \textbf{Time (s / kimg)} & \textbf{CPU memory (G)} & \textbf{GPU memory (G)} \\
\midrule
StyleGAN2 & \textbf{9.73} & \textbf{5.354} & 9.127  \\
\rowcolor{mygray}+ MS$^3$D & 9.75 & 5.374 & 9.127  \\
\bottomrule
\end{tabular}
}
\end{sc}
\end{small}
\end{center}
\vspace{-0.2in}
\end{table}

\subsection{Analysis and Ablation Studies}

\textbf{Training dynamics.}
In Section~\ref{sec:perceptual_narrowing}, we demonstrate that perceptual narrowing forms a feedback system $\Psi(x;\phi)$ characterized by high learnability and sensitivity, exhibiting significant self-dissimilarity (SD). Thus, we define multi-scale structural self-dissimilarity (MS$^3$D) and incorporate it to enhance the performance of GANs with limited data. To validate the efficacy of MS$^3$D regularization ($\mathcal{D}_\Gamma$), we illustrate the relationship between perceptual narrowing and MS$^3$D in Fig.~\ref{fig:ratio_down}(a), revealing their close connection. Fig.~\ref{fig:ratio_down}(b) displays the gradient patterns post MS$^3$D application, showing a more dispersed pattern similar to augmentation methods, which is visualized in Fig.~\ref{fig:ratio_down}(c). In Fig.~\ref{fig:ratio_down}(d), validation data demonstrate our method's superior generalization, evidenced by the discriminator's lower confidence in evaluating validation data. Moreover, with minimal training data, our method ensures stable training without collapsing. Fig.~\ref{fig:ratio_down}(e) shows that after applying MS$^3$D, the Fisher information of $\Psi(x;\phi)$ decreases, indicating more stable training dynamics.

\textbf{Loss landscape.}
In addition to examining the impact of MS$^3$D on GAN learning dynamics from the RG perspective, we also assess it from the loss landscape perspective. Figs.~\ref{fig:landscape_sim}(a) and (b) reveal how MS$^3$D influences the discriminator's loss landscape under limited data conditions, making the loss landscape flatter. A flatter loss landscape generally implies enhanced model generalization and training stability~\cite{nips/Li0TSG18}.

\textbf{Embedding space.}
Further observations show that the mean cosine similarity of the discriminator's data embeddings significantly increases by the end of training, compared to the initial phase. This increase indicates a narrower embedding space for the discriminator, suggesting a large gap between real and generated image representations. This discrepancy can enable the discriminator to easily achieve perfect true-false classification, leading to overfitting. Conversely, the application of MS$^3$D or data augmentation prevents this narrowing, maintaining low average cosine similarity levels, as depicted in Fig.~\ref{fig:landscape_sim}(c). These findings imply that perceptual narrowing may cause a narrower embedding space, and MS$^3$D can effectively mitigate this issue, thereby enhancing discriminator generalization. Additional details are available in Appendix A.

\textbf{Computational complexity.}
Our RG process involves iteratively downsampling $\Psi(x;\phi)$ and computing differences pre- and post-downsampling. Using pooling operators in PyTorch, this procedure does not require trainable parameters. Table~\ref{tab:comp_complexity} summarizes the computational overhead and memory usage during training on OxfordDog with Nvidia RTX 4090, indicating negligible resource demands.

\begin{figure}[t]
\setlength{\tabcolsep}{1pt}
\centering
\subfigure[Parameter $\lambda$]{
\includegraphics[width=0.298\linewidth]{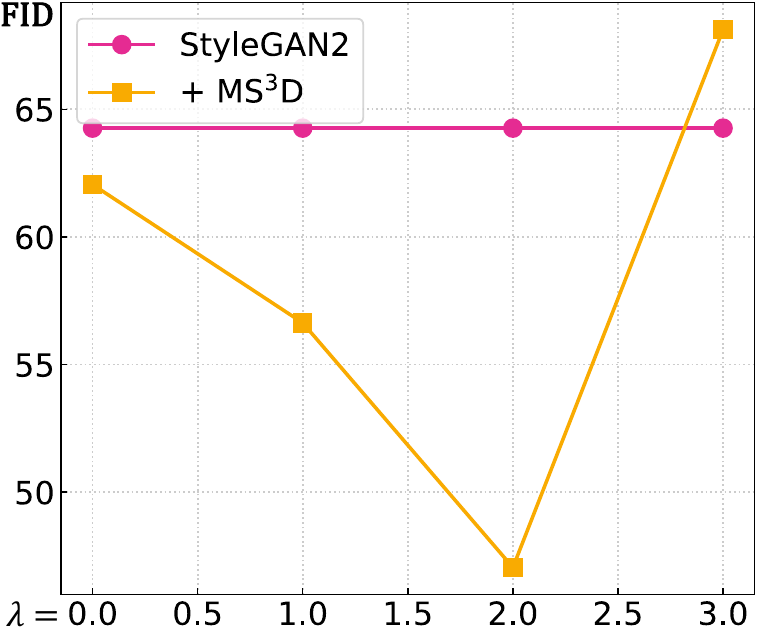}
}
\hspace{-0.08in}
\subfigure[Filter $\Gamma$]{
\includegraphics[width=0.31\linewidth]{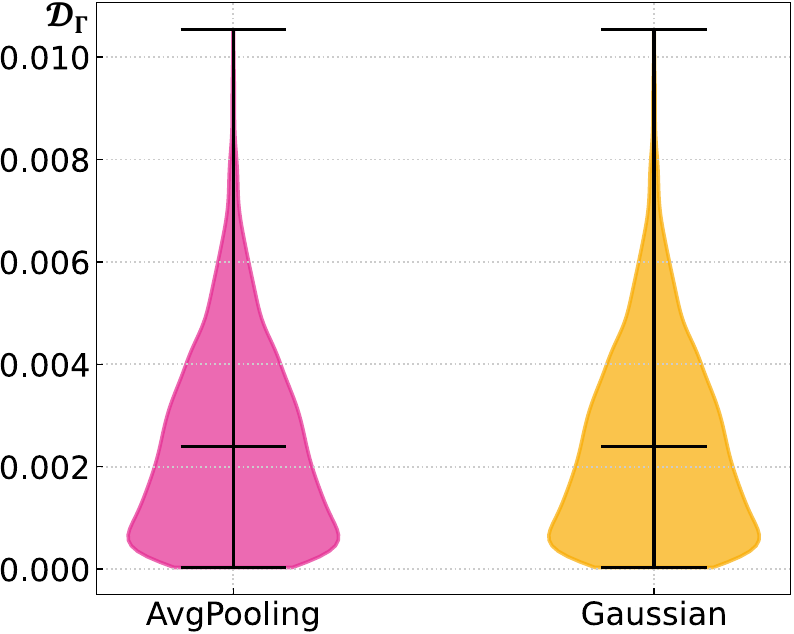}
}
\hspace{-0.08in}
\subfigure[Factor $\zeta$]{
\includegraphics[width=0.31\linewidth]{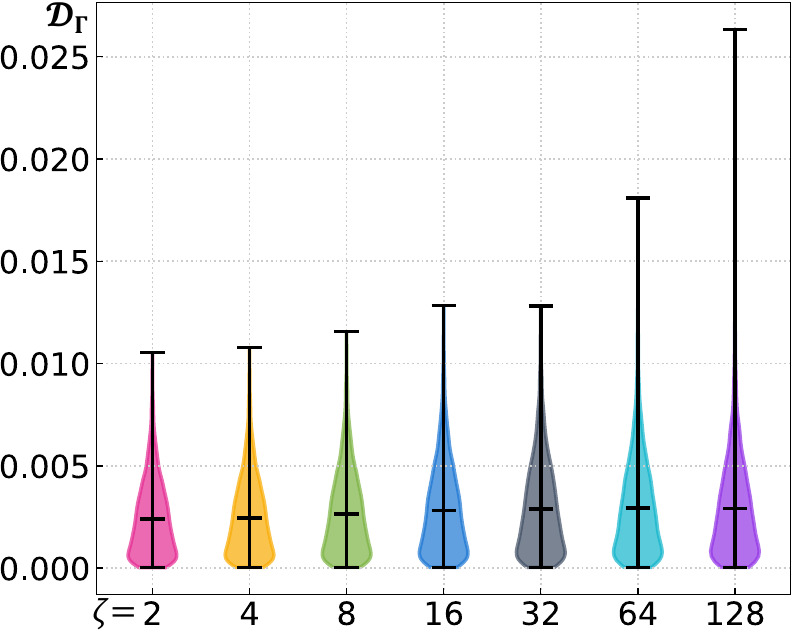}
}
\vspace{-0.1in}
\caption{Results of the ablation study. (a) Ablation study on the hyper-parameter $\lambda$. (b) Ablation of the RG transformation filter. (c) Ablation of the coarse-graining factor.}
\label{fig:ablation}
\vspace{-0.05in}
\end{figure}

\begin{table}[!tbp]
\caption{Ablation study on MS$^3$D regularization applied to real and generated data.}
\label{tab:ablation_x}
\begin{center}
\begin{sc}
\resizebox*{0.8\linewidth}{!}{
\begin{tabular}{cccccc}
\toprule
\multicolumn{2}{c}{\textbf{Configuration} $\mathcal{D}_\Gamma(\Psi(x))$} & \multirow{2}*{\textbf{OxfordDog}} & \multirow{2}*{\textbf{FFHQ-2.5K}} \\
\cmidrule(r){1-2}
$x=$ Real & $x=$ Generated & \\ 
\midrule
\XSolidBrush & \Checkmark & 62.40 & 86.51 \\ 
\Checkmark & \XSolidBrush & 47.05 & 33.46 \\
\rowcolor{mygray}\Checkmark & \Checkmark & \textbf{42.73} & \textbf{36.31} \\ 
\bottomrule
\end{tabular}
}
\end{sc}
\end{center}
\vspace{-0.1in}
\end{table}

\begin{figure*}[!tbp]
\setlength{\tabcolsep}{1pt}
\centering
\includegraphics[width=\linewidth]{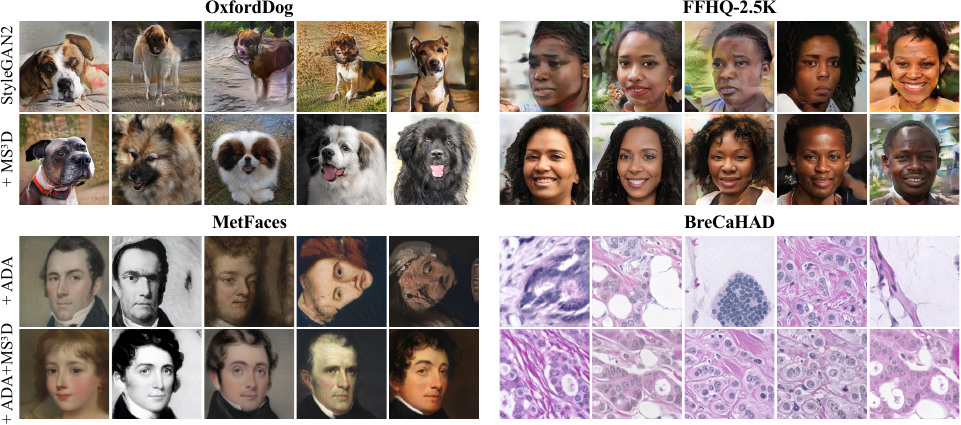}
\vspace{-0.3in}
\caption{Example generated images for several datasets with limited amount of training data.}
\label{fig:main_qualitative}
\vspace{-0.1in}
\end{figure*}

\textbf{Ablation study.}
Fig.~\ref{fig:ablation} (a) presents results for different $\lambda$ values in Eq.~(\ref{eq:loss_dis}), showing optimal performance at $\lambda = 10$. Larger $\lambda$ values may interfere with other constraint terms in StyleGAN2, which has multiple constraints.

We also conduct ablation experiments on different RG transformations, introducing a Gaussian kernel as an alternative to the Kadanoff block-spin transformation (Fig.~\ref{fig:ablation}(b)).
The results show minimal differences, leading us to adopt the simpler method. In Eq.~(\ref{eq:coarse_grain}), we chose a coarse-graining factor of 2, a common practice, and varied this factor in numerical experiments on OxfordDog (Fig.~\ref{fig:ablation}(c)), noting that larger factors increased variance and reduced numerical stability.

For $\mathcal{D}_{\Gamma_{0\to t}}(\Psi(x;\phi))$ in Eq.~(\ref{eq:loss_dis}), $x$ can be either real or generated data. Table~\ref{tab:ablation_x} compares results on OxfordDog and FFHQ-2.5K for real and generated data, finding optimal performance when $x$ includes both. However, regularization on real data alone is more effective in avoiding discriminator overfitting, leading us to choose this for computational efficiency.

\begin{table}
\caption{Quantitative results on SNGAN with FFHQ-2.5K.}
\label{tab:limitation}
\begin{center}
\begin{small}
\begin{sc}
\resizebox*{0.19\textwidth}{!}{
\begin{tabular}{lrr}
\toprule
\textbf{Method} & \textbf{FID} $\downarrow$ & \textbf{IS} $\uparrow$ \\
\midrule
SNGAN & \textbf{45.31} & 4.32  \\
\rowcolor{mygray}+ MS$^3$D & 46.14 & \textbf{4.37}  \\
\bottomrule
\end{tabular}
}
\end{sc}
\end{small}
\end{center}
\vspace{-0.2in}
\end{table}

\textbf{Limitations.}
For low-resolution image generation models (e.g., 32$\times$32), such as DCGAN and SNGAN, our method may not yield significant improvements. These models often exhibit stable training dynamics due to their reduced internal differences. Experiments conducted on SNGAN with FFHQ-2.5K, which intentionally minimizes internal differences (details in Appendix C), support this observation. The results, presented in Table~\ref{tab:limitation}, confirm that GANs trained on such datasets tend to be relatively easy to stabilize. Furthermore, the influence of our method on gradients, which promotes more uniform and dispersed gradients, may not be particularly beneficial for low-resolution image generation.

\section{Conclusion and Future Work}
In our study, we are analyzing the gradient pattern that the generator receives from the discriminator at various scales based RG flow. In particular, we are interested in the relationship between the gradient pattern and discriminator deterioration in the limited data setting.
Based on this, we propose a new regularization method, called MS$^3$D. We conduct extensive experiments on various datasets and tasks, and the results show that our method can significantly improve the performance of GANs under limited data.
Recently, RG theory has been linked with the notions of information theory~\cite{prl/gordon2021}. Therefore, an intriguing future direction is to explore the connection between the gradient pattern and the perceptual narrowing phenomenon from the angle of information theory.

\section*{Acknowledgements}
This work is supported by the Fundamental Research Funds for the Central Universities under Grant 1082204112364.

\section*{Impact Statement}
This paper presents work whose goal is to advance the field of Machine Learning. There are many potential societal consequences of our work, none which we feel must be specifically highlighted here.

\nocite{langley00}

\bibliography{references}

\begin{thebibliography}{70}
\providecommand{\natexlab}[1]{#1}
\providecommand{\url}[1]{\texttt{#1}}
\expandafter\ifx\csname urlstyle\endcsname\relax
  \providecommand{\doi}[1]{doi: #1}\else
  \providecommand{\doi}{doi: \begingroup \urlstyle{rm}\Url}\fi

\bibitem[Abadi et~al.(2016)Abadi, Agarwal, Barham, Brevdo, Chen, Citro, Corrado, Davis, Dean, Devin, Ghemawat, Goodfellow, Harp, Irving, Isard, Jia, J{\'{o}}zefowicz, Kaiser, Kudlur, Levenberg, Man{\'{e}}, Monga, Moore, Murray, Olah, Schuster, Shlens, Steiner, Sutskever, Talwar, Tucker, Vanhoucke, Vasudevan, Vi{\'{e}}gas, Vinyals, Warden, Wattenberg, Wicke, Yu, and Zheng]{corr/AbadiABBCCCDDDG16}
Abadi, M., Agarwal, A., Barham, P., Brevdo, E., Chen, Z., Citro, C., Corrado, G.~S., Davis, A., Dean, J., Devin, M., Ghemawat, S., Goodfellow, I.~J., Harp, A., Irving, G., Isard, M., Jia, Y., J{\'{o}}zefowicz, R., Kaiser, L., Kudlur, M., Levenberg, J., Man{\'{e}}, D., Monga, R., Moore, S., Murray, D.~G., Olah, C., Schuster, M., Shlens, J., Steiner, B., Sutskever, I., Talwar, K., Tucker, P.~A., Vanhoucke, V., Vasudevan, V., Vi{\'{e}}gas, F.~B., Vinyals, O., Warden, P., Wattenberg, M., Wicke, M., Yu, Y., and Zheng, X.
\newblock Tensorflow: Large-scale machine learning on heterogeneous distributed systems.
\newblock \emph{CoRR}, abs/1603.04467, 2016.

\bibitem[Achille et~al.(2019)Achille, Rovere, and Soatto]{iclr/AchilleRS19}
Achille, A., Rovere, M., and Soatto, S.
\newblock Critical learning periods in deep networks.
\newblock In \emph{7th International Conference on Learning Representations}, 2019.

\bibitem[Aksac et~al.(2019)Aksac, Demetrick, Ozyer, and Alhajj]{brn/Aksac2019}
Aksac, A., Demetrick, D.~J., Ozyer, T., and Alhajj, R.
\newblock Brecahad: a dataset for breast cancer histopathological annotation and diagnosis.
\newblock \emph{BMC Research Notes}, Dec 2019.

\bibitem[Amari(2016)]{amari2016information}
Amari, S.-i.
\newblock \emph{Information geometry and its applications}, volume 194.
\newblock Springer, 2016.

\bibitem[Arjovsky \& Bottou(2017)Arjovsky and Bottou]{iclr/ArjovskyB17}
Arjovsky, M. and Bottou, L.
\newblock Towards principled methods for training generative adversarial networks.
\newblock In \emph{5th International Conference on Learning Representations}, 2017.

\bibitem[Arjovsky et~al.(2017)Arjovsky, Chintala, and Bottou]{icml/ArjovskyCB17}
Arjovsky, M., Chintala, S., and Bottou, L.
\newblock Wasserstein generative adversarial networks.
\newblock In \emph{Proceedings of the 34th International Conference on Machine Learning}, volume~70, pp.\  214--223, 2017.

\bibitem[Bagrov et~al.(2020)Bagrov, Iakovlev, Iliasov, Katsnelson, and Mazurenko]{naos/bagrov2020}
Bagrov, A.~A., Iakovlev, I.~A., Iliasov, A.~A., Katsnelson, M.~I., and Mazurenko, V.~V.
\newblock Multiscale structural complexity of natural patterns.
\newblock \emph{Proceedings of the National Academy of Sciences}, 117\penalty0 (48):\penalty0 30241--30251, 2020.

\bibitem[Binkowski et~al.(2018)Binkowski, Sutherland, Arbel, and Gretton]{iclr/BinkowskiSAG18}
Binkowski, M., Sutherland, D.~J., Arbel, M., and Gretton, A.
\newblock Demystifying {MMD} gans.
\newblock In \emph{6th International Conference on Learning Representations}, 2018.

\bibitem[Bradde \& Bialek(2017)Bradde and Bialek]{josp/bradde2017}
Bradde, S. and Bialek, W.
\newblock Pca meets rg.
\newblock \emph{Journal of statistical physics}, 167:\penalty0 462--475, 2017.

\bibitem[Brock et~al.(2019)Brock, Donahue, and Simonyan]{iclr/BrockDS19}
Brock, A., Donahue, J., and Simonyan, K.
\newblock Large scale {GAN} training for high fidelity natural image synthesis.
\newblock In \emph{7th International Conference on Learning Representations}, 2019.

\bibitem[Cam(1986)]{ssis/Cam1986}
Cam, L.~L.
\newblock \emph{Asymptotic Methods in Statistical Decision Theory}.
\newblock Springer, Jan 1986.

\bibitem[Chen et~al.(2019)Chen, Zhai, Ritter, Lucic, and Houlsby]{cvpr/ChenZRLH19}
Chen, T., Zhai, X., Ritter, M., Lucic, M., and Houlsby, N.
\newblock Self-supervised gans via auxiliary rotation loss.
\newblock In \emph{{IEEE} Conference on Computer Vision and Pattern Recognition}, pp.\  12154--12163, 2019.

\bibitem[Cui et~al.(2022)Cui, Huang, Luo, Zhang, Zhan, and Lu]{aaai/Cui0LZZL22}
Cui, K., Huang, J., Luo, Z., Zhang, G., Zhan, F., and Lu, S.
\newblock Genco: Generative co-training for generative adversarial networks with limited data.
\newblock In \emph{Thirty-Sixth {AAAI} Conference on Artificial Intelligence}, pp.\  499--507, 2022.

\bibitem[Cui et~al.(2023)Cui, Yu, Zhan, Liao, Lu, and Xing]{cvpr/CuiYZLLX23}
Cui, K., Yu, Y., Zhan, F., Liao, S., Lu, S., and Xing, E.~P.
\newblock {KD-DLGAN:} data limited image generation via knowledge distillation.
\newblock In \emph{{IEEE/CVF} Conference on Computer Vision and Pattern Recognition}, pp.\  3872--3882, 2023.

\bibitem[Denton et~al.(2015)Denton, Chintala, Szlam, and Fergus]{nips/DentonCSF15}
Denton, E.~L., Chintala, S., Szlam, A., and Fergus, R.
\newblock Deep generative image models using a laplacian pyramid of adversarial networks.
\newblock In \emph{Advances in Neural Information Processing Systems 28: Annual Conference on Neural Information Processing Systems 2015}, pp.\  1486--1494, 2015.

\bibitem[Fang et~al.(2022)Fang, Sun, and Schwing]{nips/Fang0S22}
Fang, T., Sun, R., and Schwing, A.~G.
\newblock Diggan: Discriminator gradient gap regularization for {GAN} training with limited data.
\newblock In \emph{Advances in Neural Information Processing Systems 35: Annual Conference on Neural Information Processing Systems 2022}, volume~35, pp.\  31782--31795, 2022.

\bibitem[Fisher(1998)]{romp/fisher1998}
Fisher, M.~E.
\newblock Renormalization group theory: Its basis and formulation in statistical physics.
\newblock \emph{Reviews of Modern Physics}, 70\penalty0 (2):\penalty0 653, 1998.

\bibitem[Goodfellow et~al.(2014)Goodfellow, Pouget{-}Abadie, Mirza, Xu, Warde{-}Farley, Ozair, Courville, and Bengio]{nips/GoodfellowPMXWOCB14}
Goodfellow, I.~J., Pouget{-}Abadie, J., Mirza, M., Xu, B., Warde{-}Farley, D., Ozair, S., Courville, A.~C., and Bengio, Y.
\newblock Generative adversarial nets.
\newblock In \emph{Advances in Neural Information Processing Systems 27: Annual Conference on Neural Information Processing Systems 2014}, pp.\  2672--2680, 2014.

\bibitem[Gordon et~al.(2021)Gordon, Banerjee, Koch-Janusz, and Ringel]{prl/gordon2021}
Gordon, A., Banerjee, A., Koch-Janusz, M., and Ringel, Z.
\newblock Relevance in the renormalization group and in information theory.
\newblock \emph{Physical Review Letters}, 126\penalty0 (24):\penalty0 240601, 2021.

\bibitem[Gulrajani et~al.(2017)Gulrajani, Ahmed, Arjovsky, Dumoulin, and Courville]{nips/GulrajaniAADC17}
Gulrajani, I., Ahmed, F., Arjovsky, M., Dumoulin, V., and Courville, A.~C.
\newblock Improved training of wasserstein gans.
\newblock In \emph{Advances in Neural Information Processing Systems 30: Annual Conference on Neural Information Processing Systems 2017}, pp.\  5767--5777, 2017.

\bibitem[Gulrajani et~al.(2019)Gulrajani, Raffel, and Metz]{iclr/GulrajaniRM19}
Gulrajani, I., Raffel, C., and Metz, L.
\newblock Towards {GAN} benchmarks which require generalization.
\newblock In \emph{7th International Conference on Learning Representations}, 2019.

\bibitem[Hensch(2004)]{aron/Hensch2004}
Hensch, T.~K.
\newblock Critical period regulation.
\newblock \emph{Annual Review of Neuroscience}, pp.\  549--579, Jul 2004.

\bibitem[Heusel et~al.(2017)Heusel, Ramsauer, Unterthiner, Nessler, and Hochreiter]{nips/HeuselRUNH17}
Heusel, M., Ramsauer, H., Unterthiner, T., Nessler, B., and Hochreiter, S.
\newblock Gans trained by a two time-scale update rule converge to a local nash equilibrium.
\newblock In \emph{Advances in Neural Information Processing Systems 30: Annual Conference on Neural Information Processing Systems 2017}, pp.\  6626--6637, 2017.

\bibitem[Jacobs \& Jacobs(1992)Jacobs and Jacobs]{ds/Jacobs1992}
Jacobs, K. and Jacobs, K.
\newblock Elements of information theory.
\newblock \emph{Discrete Stochastics}, pp.\  155--183, 1992.

\bibitem[Jenni \& Favaro(2019)Jenni and Favaro]{cvpr/JenniF19}
Jenni, S. and Favaro, P.
\newblock On stabilizing generative adversarial training with noise.
\newblock In \emph{{IEEE} Conference on Computer Vision and Pattern Recognition}, pp.\  12145--12153, 2019.

\bibitem[Jiang et~al.(2021)Jiang, Dai, Wu, and Loy]{nips/JiangDWL21}
Jiang, L., Dai, B., Wu, W., and Loy, C.~C.
\newblock Deceive {D:} adaptive pseudo augmentation for {GAN} training with limited data.
\newblock In \emph{Advances in Neural Information Processing Systems 34: Annual Conference on Neural Information Processing Systems 2021}, pp.\  21655--21667, 2021.

\bibitem[Jolicoeur{-}Martineau(2019)]{iclr/Jolicoeur-Martineau19}
Jolicoeur{-}Martineau, A.
\newblock The relativistic discriminator: a key element missing from standard {GAN}.
\newblock In \emph{7th International Conference on Learning Representations}, 2019.

\bibitem[Kadanoff(1966)]{ppf/kadanoff1966}
Kadanoff, L.~P.
\newblock Scaling laws for ising models near t c.
\newblock \emph{Physics Physique Fizika}, 2\penalty0 (6):\penalty0 263, 1966.

\bibitem[Karras et~al.(2018)Karras, Aila, Laine, and Lehtinen]{iclr/KarrasALL18}
Karras, T., Aila, T., Laine, S., and Lehtinen, J.
\newblock Progressive growing of gans for improved quality, stability, and variation.
\newblock In \emph{6th International Conference on Learning Representations}, 2018.

\bibitem[Karras et~al.(2019)Karras, Laine, and Aila]{cvpr/KarrasLA19}
Karras, T., Laine, S., and Aila, T.
\newblock A style-based generator architecture for generative adversarial networks.
\newblock In \emph{{IEEE} Conference on Computer Vision and Pattern Recognition}, pp.\  4401--4410, 2019.

\bibitem[Karras et~al.(2020{\natexlab{a}})Karras, Aittala, Hellsten, Laine, Lehtinen, and Aila]{nips/KarrasAHLLA20}
Karras, T., Aittala, M., Hellsten, J., Laine, S., Lehtinen, J., and Aila, T.
\newblock Training generative adversarial networks with limited data.
\newblock In \emph{Advances in Neural Information Processing Systems 33: Annual Conference on Neural Information Processing Systems 2020}, 2020{\natexlab{a}}.

\bibitem[Karras et~al.(2020{\natexlab{b}})Karras, Laine, Aittala, Hellsten, Lehtinen, and Aila]{cvpr/KarrasLAHLA20}
Karras, T., Laine, S., Aittala, M., Hellsten, J., Lehtinen, J., and Aila, T.
\newblock Analyzing and improving the image quality of stylegan.
\newblock In \emph{2020 {IEEE/CVF} Conference on Computer Vision and Pattern Recognition}, pp.\  8107--8116, 2020{\natexlab{b}}.

\bibitem[Kirkpatrick et~al.(2016)Kirkpatrick, Pascanu, Rabinowitz, Veness, Desjardins, Rusu, Milan, Quan, Ramalho, Grabska{-}Barwinska, Hassabis, Clopath, Kumaran, and Hadsell]{corr/KirkpatrickPRVD16}
Kirkpatrick, J., Pascanu, R., Rabinowitz, N.~C., Veness, J., Desjardins, G., Rusu, A.~A., Milan, K., Quan, J., Ramalho, T., Grabska{-}Barwinska, A., Hassabis, D., Clopath, C., Kumaran, D., and Hadsell, R.
\newblock Overcoming catastrophic forgetting in neural networks.
\newblock \emph{CoRR}, abs/1612.00796, 2016.

\bibitem[Koch-Janusz \& Ringel(2018)Koch-Janusz and Ringel]{np/koch2018}
Koch-Janusz, M. and Ringel, Z.
\newblock Mutual information, neural networks and the renormalization group.
\newblock \emph{Nature Physics}, 14\penalty0 (6):\penalty0 578--582, 2018.

\bibitem[Kumari et~al.(2022)Kumari, Zhang, Shechtman, and Zhu]{cvpr/Kumari0SZ22}
Kumari, N., Zhang, R., Shechtman, E., and Zhu, J.
\newblock Ensembling off-the-shelf models for {GAN} training.
\newblock In \emph{{IEEE/CVF} Conference on Computer Vision and Pattern Recognition}, pp.\  10641--10652, 2022.

\bibitem[Li et~al.(2018)Li, Xu, Taylor, Studer, and Goldstein]{nips/Li0TSG18}
Li, H., Xu, Z., Taylor, G., Studer, C., and Goldstein, T.
\newblock Visualizing the loss landscape of neural nets.
\newblock In \emph{Advances in Neural Information Processing Systems 31: Annual Conference on Neural Information Processing Systems 2018}, pp.\  6391--6401, 2018.

\bibitem[Lin et~al.(2017)Lin, Tegmark, and Rolnick]{josp/lin2017}
Lin, H.~W., Tegmark, M., and Rolnick, D.
\newblock Why does deep and cheap learning work so well.
\newblock \emph{Journal of Statistical Physics}, 168\penalty0 (6):\penalty0 1223–1247, Sep 2017.

\bibitem[Liu et~al.(2020)Liu, Wang, Bau, Zhu, and Torralba]{cvpr/Liu0BZ020}
Liu, S., Wang, T., Bau, D., Zhu, J., and Torralba, A.
\newblock Diverse image generation via self-conditioned gans.
\newblock In \emph{2020 {IEEE/CVF} Conference on Computer Vision and Pattern Recognition}, pp.\  14274--14283, 2020.

\bibitem[Machacek \& Vaughn(1985)Machacek and Vaughn]{npb/machacek1985}
Machacek, M.~E. and Vaughn, M.~T.
\newblock Two-loop renormalization group equations in a general quantum field theory:(iii). scalar quartic couplings.
\newblock \emph{Nuclear Physics B}, 249\penalty0 (1):\penalty0 70--92, 1985.

\bibitem[Mao et~al.(2017)Mao, Li, Xie, Lau, Wang, and Smolley]{iccv/MaoLXLWS17}
Mao, X., Li, Q., Xie, H., Lau, R. Y.~K., Wang, Z., and Smolley, S.~P.
\newblock Least squares generative adversarial networks.
\newblock In \emph{{IEEE} International Conference on Computer Vision}, pp.\  2813--2821, 2017.

\bibitem[Mehta \& Schwab(2014)Mehta and Schwab]{corr/MehtaS14}
Mehta, P. and Schwab, D.~J.
\newblock An exact mapping between the variational renormalization group and deep learning.
\newblock \emph{CoRR}, abs/1410.3831, 2014.

\bibitem[Mescheder et~al.(2018)Mescheder, Geiger, and Nowozin]{icml/MeschederGN18}
Mescheder, L.~M., Geiger, A., and Nowozin, S.
\newblock Which training methods for gans do actually converge?
\newblock In \emph{Proceedings of the 35th International Conference on Machine Learning}, volume~80, pp.\  3478--3487, 2018.

\bibitem[Miyato \& Koyama(2018)Miyato and Koyama]{iclr/MiyatoK18}
Miyato, T. and Koyama, M.
\newblock cgans with projection discriminator.
\newblock In \emph{6th International Conference on Learning Representations}, 2018.

\bibitem[Miyato et~al.(2018)Miyato, Kataoka, Koyama, and Yoshida]{iclr/MiyatoKKY18}
Miyato, T., Kataoka, T., Koyama, M., and Yoshida, Y.
\newblock Spectral normalization for generative adversarial networks.
\newblock In \emph{6th International Conference on Learning Representations}, 2018.

\bibitem[Mo et~al.(2020)Mo, Cho, and Shin]{corr/Sangwoo2020}
Mo, S., Cho, M., and Shin, J.
\newblock Freeze discriminator: {A} simple baseline for fine-tuning gans.
\newblock \emph{CoRR}, abs/2002.10964, 2020.

\bibitem[Parkhi et~al.(2012)Parkhi, Vedaldi, Zisserman, and Jawahar]{cvpr/ParkhiVZJ12}
Parkhi, O.~M., Vedaldi, A., Zisserman, A., and Jawahar, C.~V.
\newblock Cats and dogs.
\newblock In \emph{2012 {IEEE} Conference on Computer Vision and Pattern Recognition}, pp.\  3498--3505, 2012.

\bibitem[Paszke et~al.(2019)Paszke, Gross, Massa, Lerer, Bradbury, Chanan, Killeen, Lin, Gimelshein, Antiga, Desmaison, K{\"{o}}pf, Yang, DeVito, Raison, Tejani, Chilamkurthy, Steiner, Fang, Bai, and Chintala]{nips/PaszkeGMLBCKLGA19}
Paszke, A., Gross, S., Massa, F., Lerer, A., Bradbury, J., Chanan, G., Killeen, T., Lin, Z., Gimelshein, N., Antiga, L., Desmaison, A., K{\"{o}}pf, A., Yang, E.~Z., DeVito, Z., Raison, M., Tejani, A., Chilamkurthy, S., Steiner, B., Fang, L., Bai, J., and Chintala, S.
\newblock Pytorch: An imperative style, high-performance deep learning library.
\newblock In \emph{Advances in Neural Information Processing Systems 32: Annual Conference on Neural Information Processing Systems 2019}, pp.\  8024--8035, 2019.

\bibitem[Radford et~al.(2016)Radford, Metz, and Chintala]{corr/RadfordMC15}
Radford, A., Metz, L., and Chintala, S.
\newblock Unsupervised representation learning with deep convolutional generative adversarial networks.
\newblock In \emph{4th International Conference on Learning Representations}, 2016.

\bibitem[Radford et~al.(2021)Radford, Kim, Hallacy, Ramesh, Goh, Agarwal, Sastry, Askell, Mishkin, Clark, Krueger, and Sutskever]{icml/RadfordKHRGASAM21}
Radford, A., Kim, J.~W., Hallacy, C., Ramesh, A., Goh, G., Agarwal, S., Sastry, G., Askell, A., Mishkin, P., Clark, J., Krueger, G., and Sutskever, I.
\newblock Learning transferable visual models from natural language supervision.
\newblock In \emph{Proceedings of the 38th International Conference on Machine Learning}, volume 139, pp.\  8748--8763, 2021.

\bibitem[Salimans et~al.(2016)Salimans, Goodfellow, Zaremba, Cheung, Radford, and Chen]{nips/SalimansGZCRCC16}
Salimans, T., Goodfellow, I.~J., Zaremba, W., Cheung, V., Radford, A., and Chen, X.
\newblock Improved techniques for training gans.
\newblock In \emph{Advances in Neural Information Processing Systems 29: Annual Conference on Neural Information Processing Systems 2016}, pp.\  2226--2234, 2016.

\bibitem[Shorten \& Khoshgoftaar(2019)Shorten and Khoshgoftaar]{jbd/ShortenK19}
Shorten, C. and Khoshgoftaar, T.~M.
\newblock A survey on image data augmentation for deep learning.
\newblock \emph{J. Big Data}, 6:\penalty0 60, 2019.

\bibitem[S{\o}nderby et~al.(2017)S{\o}nderby, Caballero, Theis, Shi, and Husz{\'{a}}r]{iclr/SonderbyCTSH17}
S{\o}nderby, C.~K., Caballero, J., Theis, L., Shi, W., and Husz{\'{a}}r, F.
\newblock Amortised {MAP} inference for image super-resolution.
\newblock In \emph{5th International Conference on Learning Representations}, 2017.

\bibitem[Song \& Ermon(2020)Song and Ermon]{icml/SongE20}
Song, J. and Ermon, S.
\newblock Bridging the gap between f-gans and wasserstein gans.
\newblock In \emph{Proceedings of the 37th International Conference on Machine Learning}, volume 119, pp.\  9078--9087, 2020.

\bibitem[Srivastava et~al.(2014)Srivastava, Hinton, Krizhevsky, Sutskever, and Salakhutdinov]{jmlr/SrivastavaHKSS14}
Srivastava, N., Hinton, G.~E., Krizhevsky, A., Sutskever, I., and Salakhutdinov, R.
\newblock Dropout: a simple way to prevent neural networks from overfitting.
\newblock \emph{J. Mach. Learn. Res.}, 15\penalty0 (1):\penalty0 1929--1958, 2014.

\bibitem[Stanley(1999)]{romp/stanley1999}
Stanley, H.~E.
\newblock Scaling, universality, and renormalization: Three pillars of modern critical phenomena.
\newblock \emph{Reviews of modern physics}, 71\penalty0 (2):\penalty0 S358, 1999.

\bibitem[Tran et~al.(2020)Tran, Tran, Nguyen, Nguyen, and Cheung]{corr/abs-2006-05338}
Tran, N., Tran, V., Nguyen, N., Nguyen, T., and Cheung, N.
\newblock Towards good practices for data augmentation in {GAN} training.
\newblock \emph{CoRR}, abs/2006.05338, 2020.

\bibitem[Tran et~al.(2021)Tran, Tran, Nguyen, Nguyen, and Cheung]{tip/TranTNNC21}
Tran, N., Tran, V., Nguyen, N., Nguyen, T., and Cheung, N.
\newblock On data augmentation for {GAN} training.
\newblock \emph{{IEEE} Trans. Image Process.}, 30:\penalty0 1882--1897, 2021.

\bibitem[Tseng et~al.(2021)Tseng, Jiang, Liu, Yang, and Yang]{cvpr/TsengJL0Y21}
Tseng, H., Jiang, L., Liu, C., Yang, M., and Yang, W.
\newblock Regularizing generative adversarial networks under limited data.
\newblock In \emph{{IEEE} Conference on Computer Vision and Pattern Recognition}, pp.\  7921--7931, 2021.

\bibitem[Wang et~al.(2020)Wang, Gonzalez{-}Garcia, Berga, Herranz, Khan, and van~de Weijer]{cvpr/WangGBHK020}
Wang, Y., Gonzalez{-}Garcia, A., Berga, D., Herranz, L., Khan, F.~S., and van~de Weijer, J.
\newblock Minegan: Effective knowledge transfer from gans to target domains with few images.
\newblock In \emph{2020 {IEEE/CVF} Conference on Computer Vision and Pattern Recognition}, pp.\  9329--9338, 2020.

\bibitem[Webster et~al.(2019)Webster, Rabin, Simon, and Jurie]{cvpr/WebsterRSJ19}
Webster, R., Rabin, J., Simon, L., and Jurie, F.
\newblock Detecting overfitting of deep generative networks via latent recovery.
\newblock In \emph{{IEEE} Conference on Computer Vision and Pattern Recognition}, pp.\  11273--11282, 2019.

\bibitem[Wilson(1971)]{prb/wilson1971}
Wilson, K.~G.
\newblock Renormalization group and critical phenomena. i. renormalization group and the kadanoff scaling picture.
\newblock \emph{Physical review B}, 4\penalty0 (9):\penalty0 3174, 1971.

\bibitem[Wolpert \& Macready(2007)Wolpert and Macready]{complexity/WolpertM07}
Wolpert, D.~H. and Macready, W.~G.
\newblock Using self-dissimilarity to quantify complexity.
\newblock \emph{Complex.}, 12\penalty0 (3):\penalty0 77--85, 2007.

\bibitem[Wolpert \& Macready(2018)Wolpert and Macready]{utics/wolpert2018}
Wolpert, D.~H. and Macready, W.~G.
\newblock Self-dissimilarity: An empirically observable complexity measure.
\newblock In \emph{Unifying Themes In Complex Systems, Volume 1}, pp.\  625--644. CRC Press, 2018.

\bibitem[Zhang et~al.(2017)Zhang, Xu, and Li]{iccv/ZhangXL17}
Zhang, H., Xu, T., and Li, H.
\newblock Stackgan: Text to photo-realistic image synthesis with stacked generative adversarial networks.
\newblock In \emph{{IEEE} International Conference on Computer Vision}, pp.\  5908--5916, 2017.

\bibitem[Zhang et~al.(2019)Zhang, Goodfellow, Metaxas, and Odena]{icml/ZhangGMO19}
Zhang, H., Goodfellow, I.~J., Metaxas, D.~N., and Odena, A.
\newblock Self-attention generative adversarial networks.
\newblock In \emph{Proceedings of the 36th International Conference on Machine Learning}, volume~97, pp.\  7354--7363, 2019.

\bibitem[Zhang et~al.(2020)Zhang, Zhang, Odena, and Lee]{iclr/ZhangZOL20}
Zhang, H., Zhang, Z., Odena, A., and Lee, H.
\newblock Consistency regularization for generative adversarial networks.
\newblock In \emph{8th International Conference on Learning Representations}. OpenReview.net, 2020.

\bibitem[Zhao et~al.(2017)Zhao, Mathieu, and LeCun]{iclr/ZhaoML17}
Zhao, J.~J., Mathieu, M., and LeCun, Y.
\newblock Energy-based generative adversarial networks.
\newblock In \emph{5th International Conference on Learning Representations}, 2017.

\bibitem[Zhao et~al.(2020{\natexlab{a}})Zhao, Liu, Lin, Zhu, and Han]{nips/ZhaoLLZ020}
Zhao, S., Liu, Z., Lin, J., Zhu, J., and Han, S.
\newblock Differentiable augmentation for data-efficient {GAN} training.
\newblock In \emph{Advances in Neural Information Processing Systems 33: Annual Conference on Neural Information Processing Systems 2020}, 2020{\natexlab{a}}.

\bibitem[Zhao et~al.(2020{\natexlab{b}})Zhao, Zhang, Chen, Singh, and Zhang]{corr/abs-2006-02595}
Zhao, Z., Zhang, Z., Chen, T., Singh, S., and Zhang, H.
\newblock Image augmentations for {GAN} training.
\newblock \emph{CoRR}, abs/2006.02595, 2020{\natexlab{b}}.

\bibitem[Zhao et~al.(2021)Zhao, Singh, Lee, Zhang, Odena, and Zhang]{aaai/Zhao0LZOZ21}
Zhao, Z., Singh, S., Lee, H., Zhang, Z., Odena, A., and Zhang, H.
\newblock Improved consistency regularization for gans.
\newblock In \emph{Thirty-Fifth {AAAI} Conference on Artificial Intelligence}, pp.\  11033--11041, 2021.

\end{thebibliography}
\bibliographystyle{icml2024}

\newpage
\appendix
\onecolumn

\section{Discussion}

\subsection{Simple Proof of SD Computation}
In this section, we demonstrate the proof for computing the structural self-dissimilarity (SD) as presented in the main text. We begin by defining SD as follows:
\begin{equation}
\mathcal{D}_{\Gamma_{s\to s+ds}} =\left|\langle\Gamma_{s}|\Gamma_{s+ds}\rangle-\frac{1}{2}(\langle\Gamma_{s}|\Gamma_{s}\rangle+\langle\Gamma_{s+ds}|\Gamma_{s+ds}\rangle) \right|.
\end{equation}
In the main text, we showed that SD can be computed using the equation:
\begin{equation}
\mathcal{D}_{\Gamma_{s\to s+ds}} = \frac{1}{2}\left|(\langle\Gamma_{s}|\Gamma_{s}\rangle - \langle\Gamma_{s+ds}|\Gamma_{s+ds}\rangle) \right|.
\end{equation}
\textit{proof:} Consider the Kadanoff block-spin transformation:
\begin{equation}
\left\{\Gamma_{s+ds}(\Psi(x;\phi))\right\}(i,j)=\frac{1}{(\zeta_{s})^2}\sum_{m=0}^{\zeta_{s}-1}\sum_{n=0}^{\zeta_{s}-1}\left\{\Gamma_{s}(\Psi(x;\phi))\right\}\left (\lfloor\frac{i}{\zeta_{s}}\rfloor\cdot\zeta_{s}+m, \lfloor\frac{j}{\zeta_{s}}\rfloor\cdot\zeta_{s}+n\right ).
\end{equation}
Thus, we have:
\begin{equation}
\begin{aligned}
\langle\Gamma_{s}|\Gamma_{s+ds}\rangle &= \frac{1}{(L_{s+ds})^2}\sum_{i=0}^{L_s-1}\sum_{j=0}^{L_s-1}\Gamma_{s}\left\{(\Psi(x;\phi))\right\}(i,j)\sum_{m=0}^{\zeta_{s}-1}\sum_{n=0}^{\zeta_{s}-1}\left\{\Gamma_{s}(\Psi(x;\phi))\right\}\left (\lfloor\frac{i}{\zeta_{s}}\rfloor\cdot\zeta_{s}+m, \lfloor\frac{j}{\zeta_{s}}\rfloor\cdot\zeta_{s}+n\right ) \\
&=\frac{(\zeta_{s})^2}{(L_{s+ds})^2}\sum_{i=0}^{L_s-1}\sum_{j=0}^{L_s-1}\left (\Gamma_{s}\left\{(\Psi(x;\phi))\right\}(i,j)\right )^2 \\
&=\frac{(\zeta_{s})^2}{(L_{s+ds})^2}\cdot(L_s)^2(\Gamma_{s}(\Psi(x;\phi)))^2 \\
&= \langle\Gamma_{s}|\Gamma_{s}\rangle
\end{aligned}
\end{equation}

\subsection{Comparison with ADA}
While adaptive discriminator augmentation (ADA)~\cite{nips/KarrasAHLLA20} is an effective method for training GANs under limited data, we observe that with very small datasets, ADA increases the probability of augmentation and leads to ``augmentations leak''. For instance, faces generated on MetFaces may show rotation (see Fig.~\ref{fig:main_qualitative} and~\ref{fig:main_qualitative_full}). In Fig.~\ref{fig:ratio_ada_ours}, we compare ADA in terms of the degree of aggregation of the gradients. Our approach demonstrates better performance in avoiding aggregated patterns on small datasets like MetFaces~\cite{nips/KarrasAHLLA20} and BreCaHAD~\cite{brn/Aksac2019}.

\subsection{Cosine Similarity of the Data Embedding Space}
\begin{figure}
\centering
\begin{minipage}{.5\textwidth}
    \centering
    \includegraphics[width=0.8\linewidth]{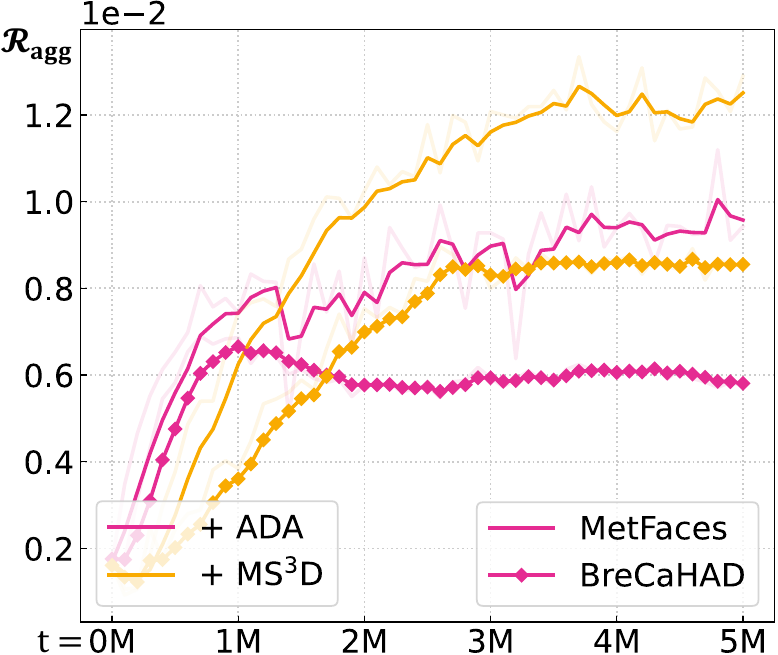}
    \caption{Degree of aggregation of the gradients.}
    \label{fig:ratio_ada_ours}
\end{minipage}%
\begin{minipage}{.5\textwidth}
    \centering
    \includegraphics[width=0.8\linewidth]{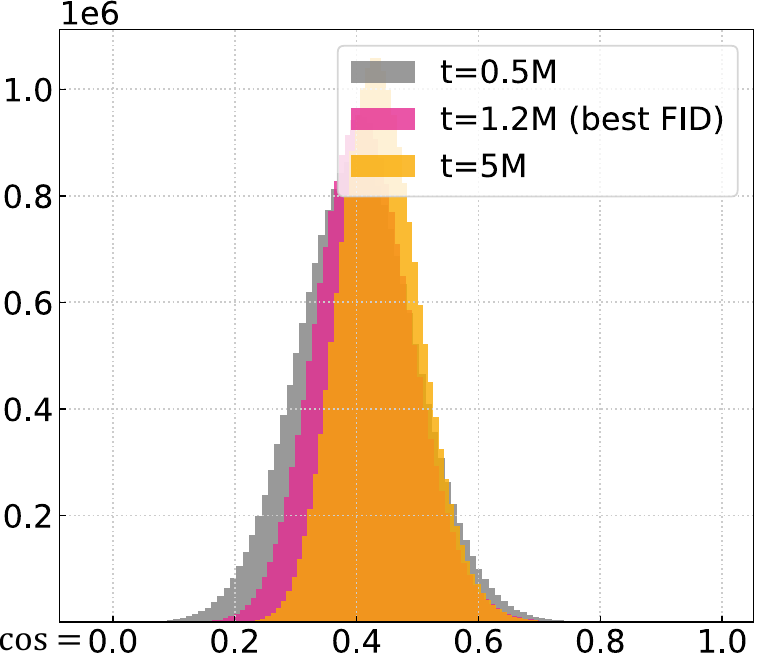}
    \caption{Cosine similarity over steps.}
    \label{fig:sim_over_steps}
\end{minipage}
\end{figure}

We use the discriminator as a feature extractor and obtain data representations through forward inference. We then compute the pairwise cosine similarity of the representations for all real images. Observing the training process, the average cosine similarity of the representations increases significantly, as shown in Fig.~\ref{fig:sim_over_steps}. A higher average cosine similarity indicates a narrowing of the representation space. Considering the surface area ratio of the unit hypersphere: in two dimensions, $\arccos(0.56)=63.90^\circ$, implying that a cosine similarity of 0.56 can "occupy"
\[
\frac{63.90^\circ}{360^\circ}=17.75\%
\]
of the 2D unit circle. In three dimensions, it is only 6.96\%. This low proportion suggests a sparse distribution of points with a cosine similarity of 0.56 in high-dimensional space, possibly explaining the narrowing of the representation space.

\subsection{Landscape of Discriminator}
\begin{figure}[t]
\setlength{\tabcolsep}{1pt}
\centering
{\small
\begin{tabular}{c c c c}
\vspace{-1pt}
\raisebox{0.095\linewidth}{\rotatebox[origin=t]{90}{StyleGAN2}}&
\includegraphics[width=0.32\linewidth]{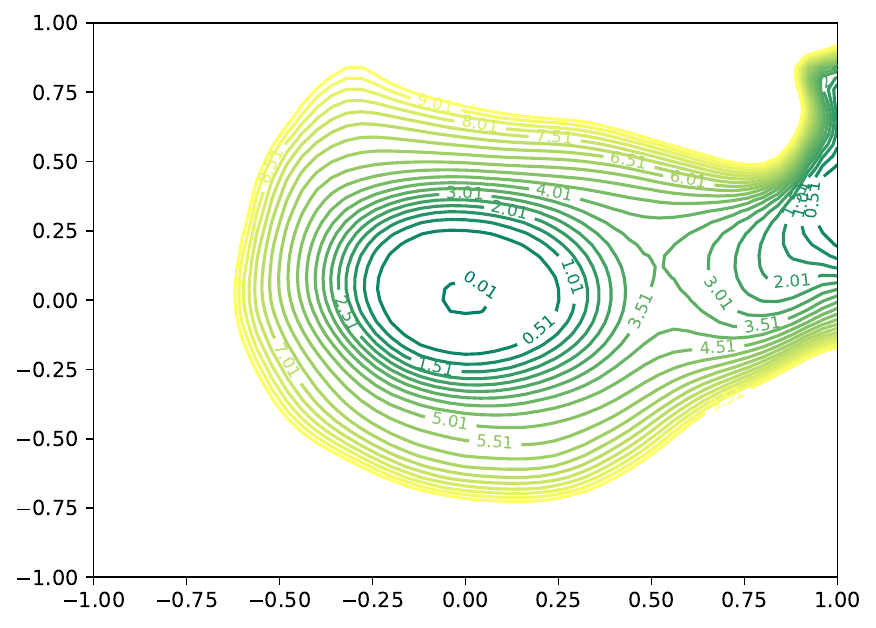}&
\includegraphics[width=0.284\linewidth]{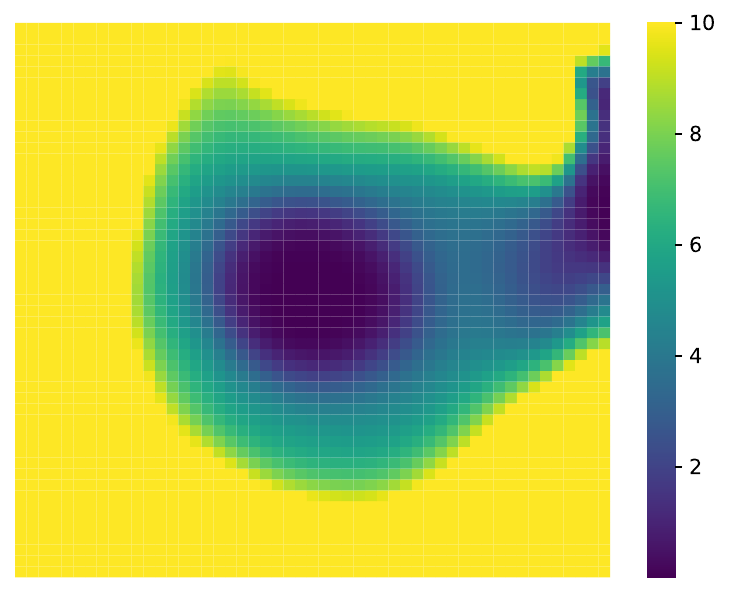}&
\includegraphics[width=0.3\linewidth]{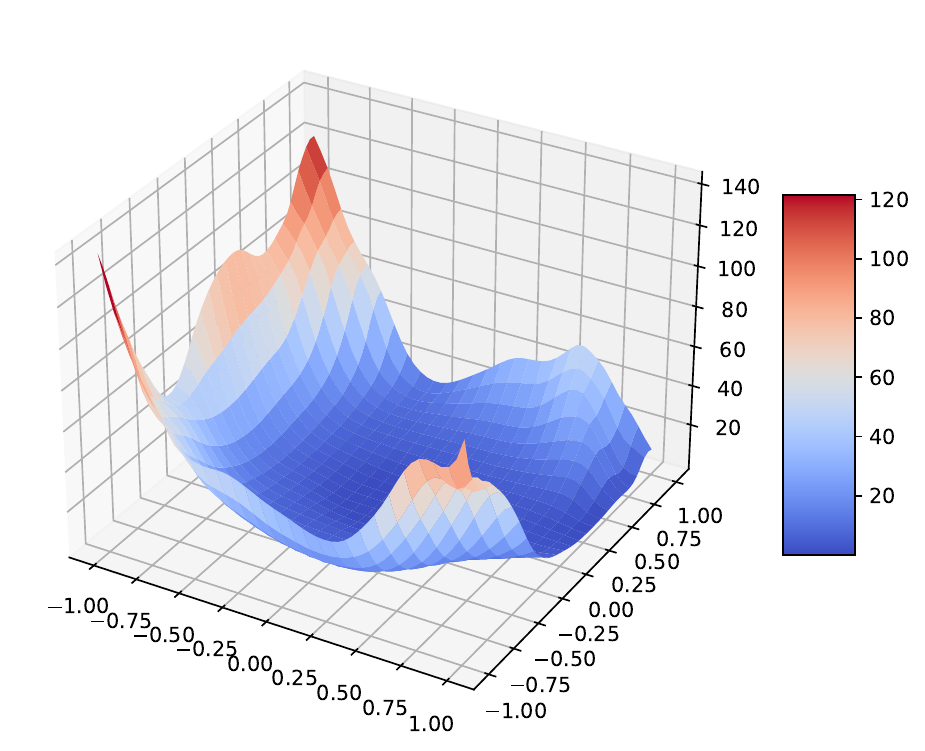}
\tabularnewline
\raisebox{0.095\linewidth}{\rotatebox[origin=t]{90}{+ ADA}}&
\includegraphics[width=0.32\linewidth]{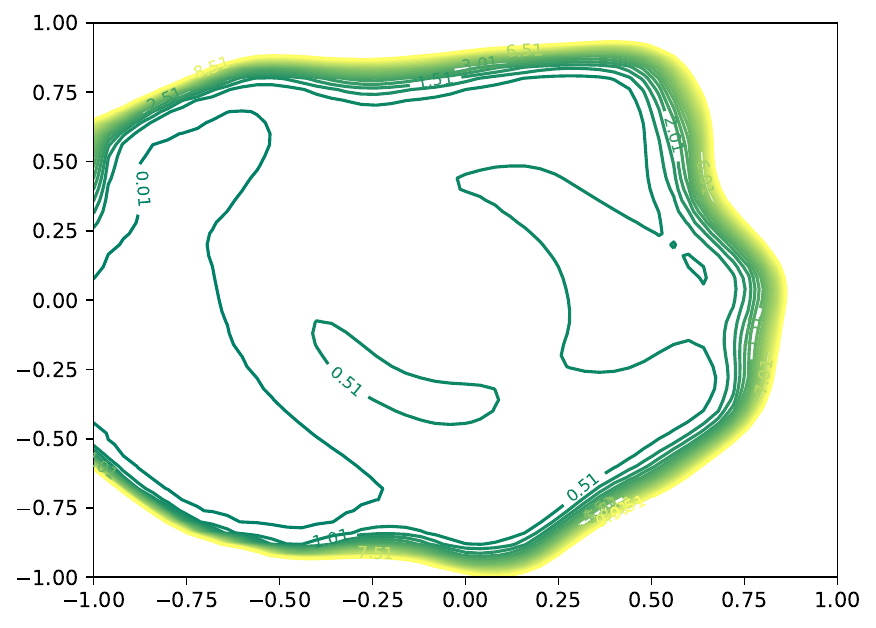}&
\includegraphics[width=0.284\linewidth]{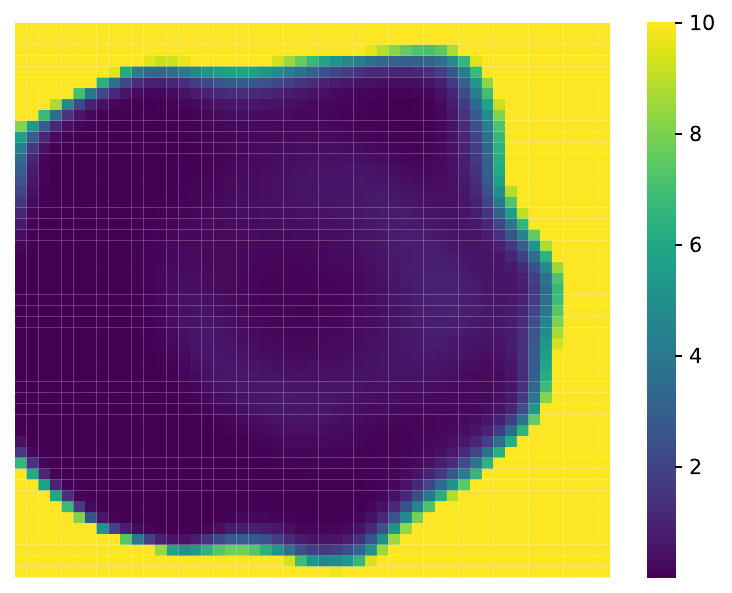}&
\includegraphics[width=0.3\linewidth]{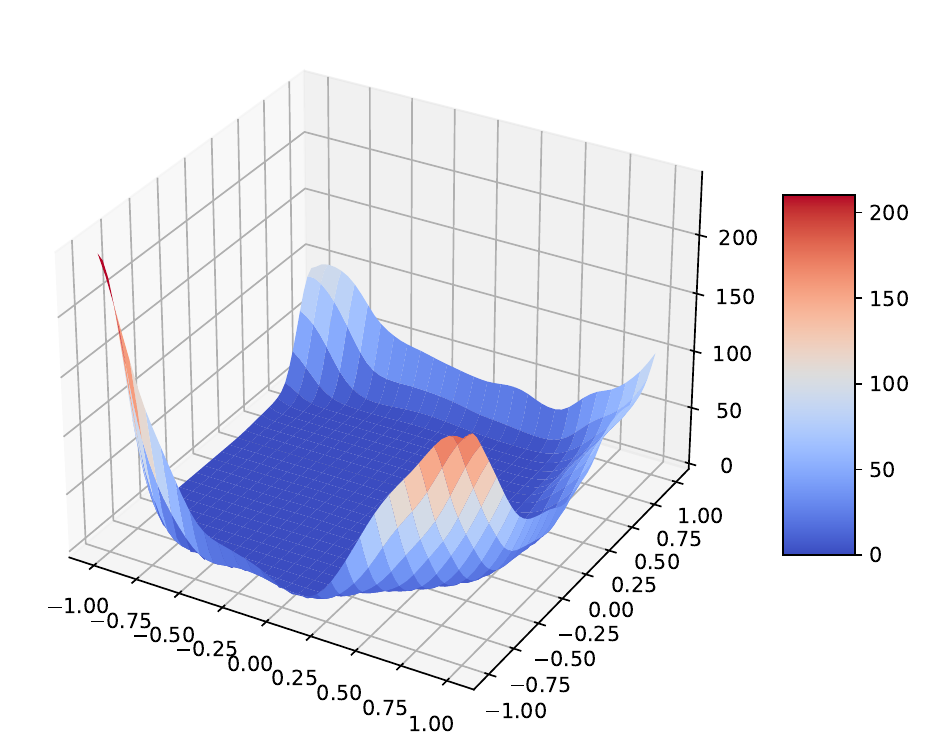}
\tabularnewline
\raisebox{0.095\linewidth}{\rotatebox[origin=t]{90}{+ MS$^3$D}}&
\includegraphics[width=0.32\linewidth]{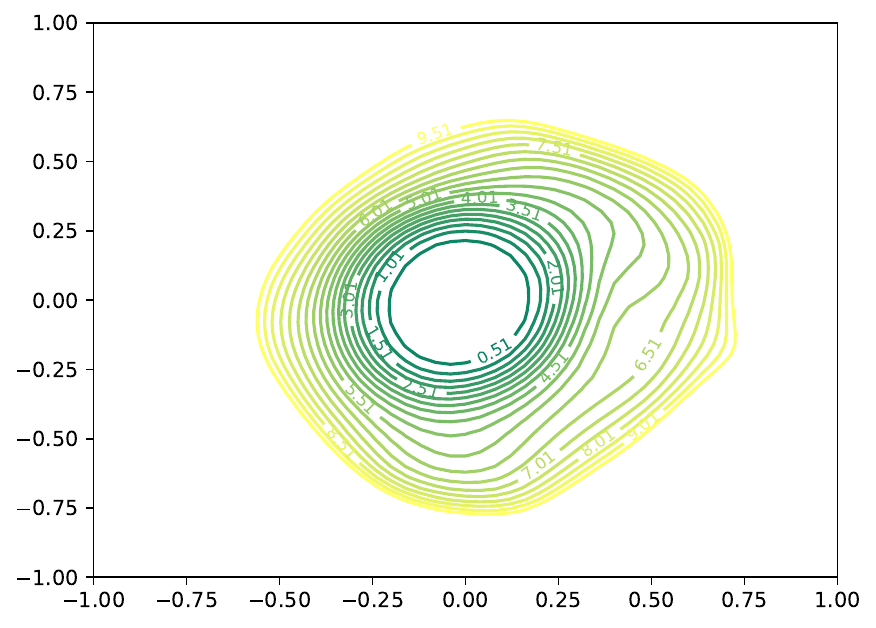}&
\includegraphics[width=0.284\linewidth]{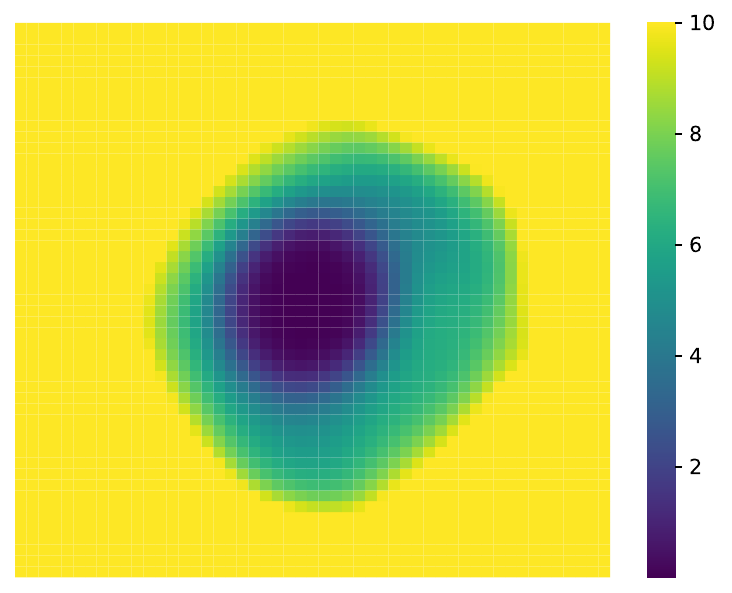}&
\includegraphics[width=0.3\linewidth]{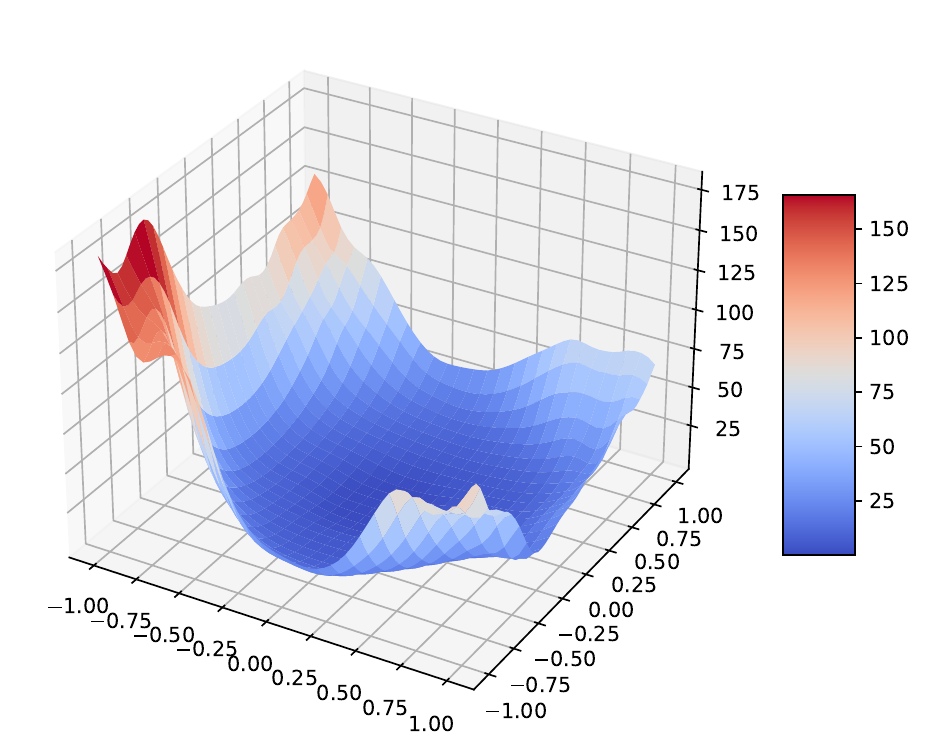}
\tabularnewline
& 2D & 2D heatmap & 3D
\end{tabular}
}
\vspace{-0.08in}
\caption{Visualization of the loss landscapes.}
\vspace{-0.2in}
\label{fig:more_landscape}
\end{figure}

In Fig.~\ref{fig:more_landscape}, we present the loss landscape of the discriminator without MS$^3$D, with MS$^3$D, and with ADA. We find that our MS$^3$D contributes to a flatter loss landscape. A flatter loss landscape often correlates with higher model generalization and training stability~\cite{nips/Li0TSG18}.

\subsection{Pattern of $\nabla_xf(x;\phi)$}

\begin{figure}
\setlength{\tabcolsep}{1pt}
\centering
\includegraphics[width=\linewidth]{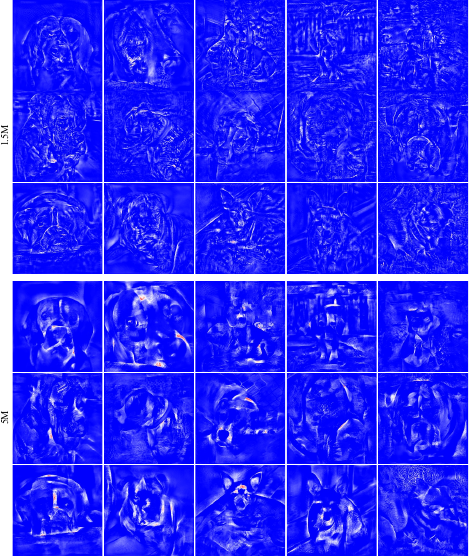}
\caption{Visualizing $\nabla_x(f(x;\phi))$ on StyleGAN2 (uncurated).}
\vspace{-0.25cm}
\label{fig:more_pattern}
\end{figure}

\begin{figure}
\setlength{\tabcolsep}{1pt}
\centering
\includegraphics[width=\linewidth]{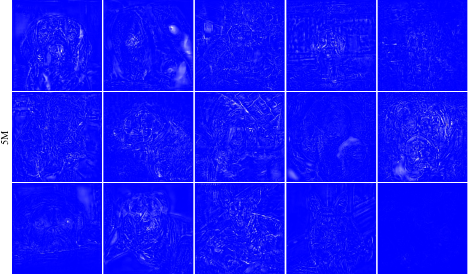}
\caption{Visualizing $\nabla_x(f(x;\phi))$ on StyleGAN2 + MS$^3$D (uncurated).}
\label{fig:more_pattern2}
\end{figure}

\begin{figure}
\setlength{\tabcolsep}{1pt}
\centering
{\small
\begin{tabular}{c c c c}
t = & 1M & 5M & 25M
\tabularnewline
\raisebox{0.05\linewidth}{\rotatebox[origin=t]{90}{\textit{w/o} ADA}}&
\includegraphics[width=0.12\linewidth]{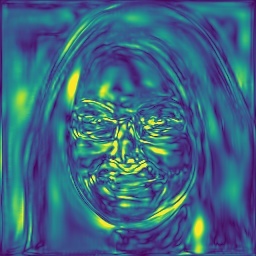}&
\includegraphics[width=0.12\linewidth]{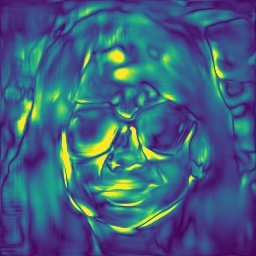}&
\includegraphics[width=0.12\linewidth]{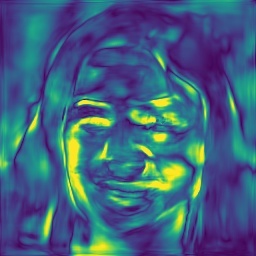}
\tabularnewline
$\mathcal{D}_\Gamma$ = & \textbf{0.0243} & \textbf{0.0256} & \textbf{0.0262}
\tabularnewline
\raisebox{0.05\linewidth}{\rotatebox[origin=t]{90}{\textit{w/} ADA}}&
\includegraphics[width=0.12\linewidth]{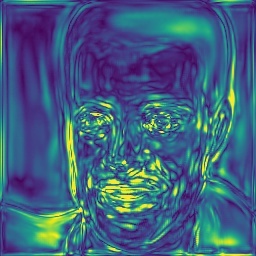}&
\includegraphics[width=0.12\linewidth]{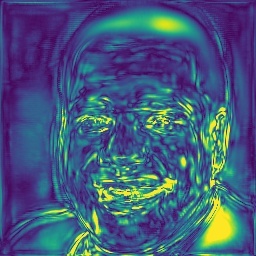}&
\includegraphics[width=0.12\linewidth]{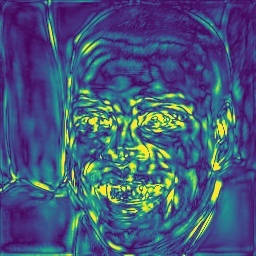}
\tabularnewline
$\mathcal{D}_\Gamma$ = & \textbf{0.0241} & \textbf{0.0245} & \textbf{0.0240}
\end{tabular}
}
\caption{MS$^3$D analysis on natural patterns. The computation of the MS$^3$D descriptor, $\mathcal{D}_\Gamma$, is performed directly on the images obtained from~\cite{nips/KarrasAHLLA20}}

\label{fig:nature_pattern_calc}
\vspace{-0.2in}
\end{figure}

Additional visualizations of $\nabla_xf(x;\phi)$ are shown in Fig.~\ref{fig:more_pattern} and~\ref{fig:more_pattern2}. In Fig.~\ref{fig:nature_pattern_calc}, we compute the multi-scale structural self-dissimilarity (MS$^3$D) of the discriminator's activation maps, revealing that MS$^3$D effectively captures the complexity of natural patterns.

\section{Illustration of MS$^3$D}

\begin{figure}[!h]
\centering
\includegraphics[width=0.9\linewidth]{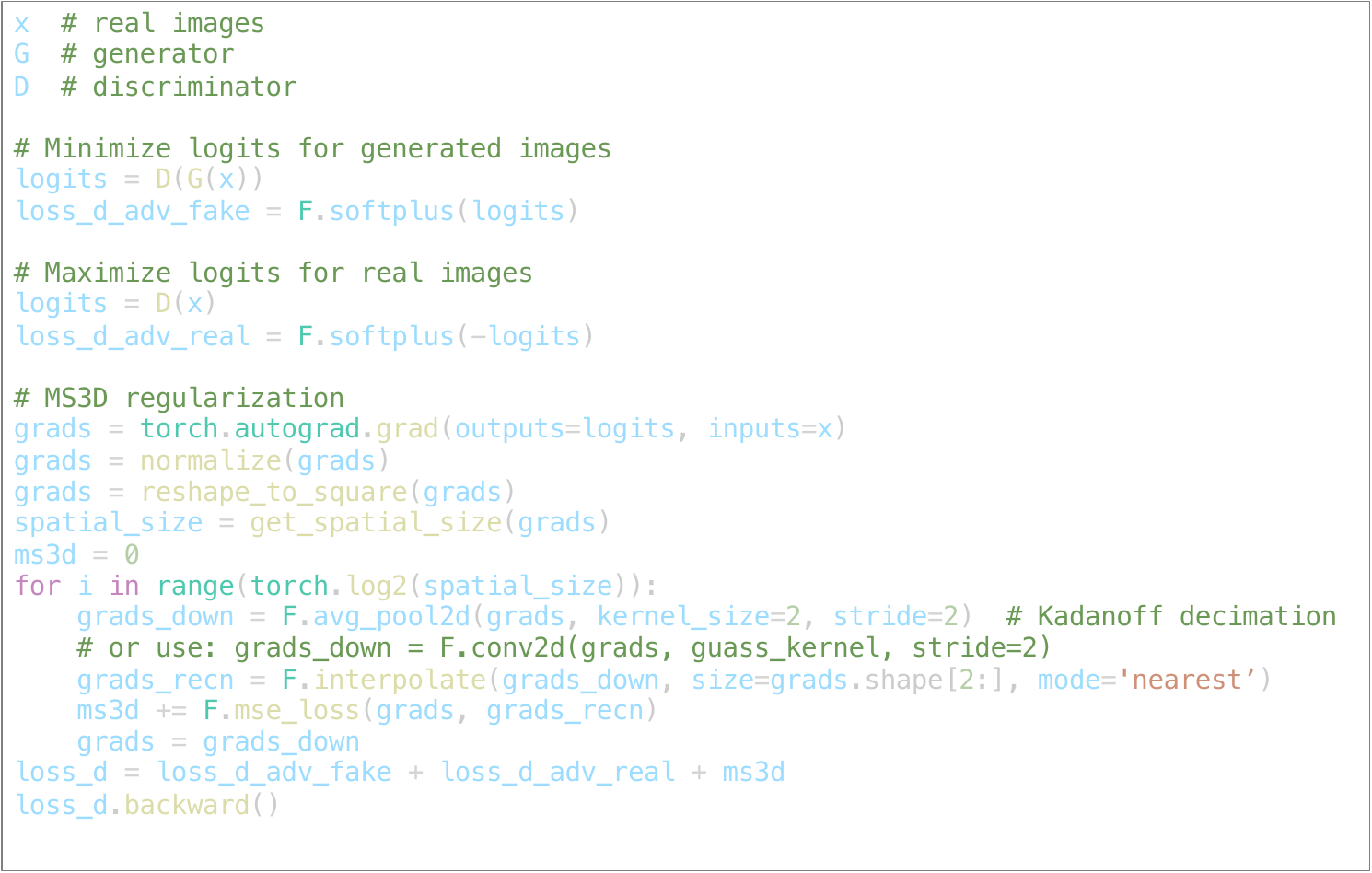}
\caption{Pseudocode of the proposed multi-scale structural self-dissimilarity (MS$^3$D) regularization.}
\label{fig:algorithm}
\vspace{-0.1in}
\end{figure}

\textbf{Pseudocode.}
In this section, we provide the pseudocode of the proposed multi-scale structural self-dissimilarity (MS$^3$D) regularization. The pseudocode is shown in Fig~\ref{fig:algorithm}.

\section{Experimental settings}
\subsection{Datasets}

\textbf{OxfordDog.}
The Oxford-IIIT pet dataset~\cite{cvpr/ParkhiVZJ12} contains images of cat and dog species. For this study, we select only dog images, apply a dog face detection model to crop and resize them to a uniform square format of 256$\times$256 pixels, centering the dog's face as much as possible. We randomly select 4,492 images for training and use the remaining 498 images as the test set. This dataset includes a wide variety of dog breeds (about 25 dog breeds), poses, and backgrounds, thus providing a challenging testbed for our experiments. We refer to this dataset as OxfordDog.

\textbf{FFHQ.}
The Flickr-Faces-HQ (FFHQ) dataset~\cite{cvpr/KarrasLA19} includes 70,000 high-quality images of human faces, covering a wide range of ages, ethnicities, and image backgrounds, as well as various accessories like eyeglasses and hats. Each image was sourced from Flickr, aligned, and cropped to ensure consistency and quality.

\textbf{MetFaces.}
The MetFaces dataset~\cite{nips/KarrasAHLLA20} comprises 1,336 high-quality facial images from the Metropolitan Museum of Art's collection \url{(https://metmuseum.github.io/)}.

\textbf{BreCaHAD.}
The BreCaHAD dataset~\cite{brn/Aksac2019}, designed for breast cancer histopathology studies, contains 162 images at 1360$\times$1024 resolution. We restructure these into 1,944 partially overlapping crops of 512x512 pixels for our experiments.

\textbf{FFHQ-2.5K.}
Using the CLIP visual encoder~\cite{icml/RadfordKHRGASAM21}, we extract features from the FFHQ images and cluster them into 14 groups using K-means. We select the smallest cluster, which contains 2,500 images (i.e., FFHQ-2.5K), for our study.

\subsection{Evaluation Metrics}
We adopt the following metrics to evaluate the effectiveness of our method and compare it with the baselines. \textbf{Inception Score (IS)}~\cite{nips/SalimansGZCRCC16} evaluates the quality and diversity of the generated images. It is a common metric in the early stage of GAN development and estimates the quality of the generated images based on the conditional entropy of the class labels predicted by an Inception network. A higher IS score indicates better performance. \textbf{Fr\'{e}chet Inception Distance (FID)}~\cite{nips/HeuselRUNH17} is another widely used metric for measuring the quality of the generated images. FID fits a Gaussian distribution to the hidden activations for each distribution (generated and ground truth) and then computes the Fr\'{e}chet distance, also known as the Wasserstein-2 distance, between those Gaussians. \textbf{Kernel Inception Distance (KID)}~\cite{iclr/BinkowskiSAG18} is a metric similar to FID, which is the squared maximum mean discrepancy (MMD) between Inception representations. Unlike FID, it does not assume that the activation distribution has a parametric form, and it has a simple unbiased estimator. It is more informative in situations where the ground truth data is small scale. A lower KID score indicates better performance.

\subsection{Setups}
We train SNGAN and StyleGAN2 using implementations from \citet{iclr/MiyatoKKY18} and \citet{cvpr/KarrasLAHLA20}, respectively, and implement a 64$\times$64 version of DCGAN following \citet{corr/RadfordMC15}. All experiments use the same hyperparameter settings, and model performance is evaluated using the implementation from \citet{cvpr/KarrasLAHLA20}. For integrating MS$^3$D with other methods, we set $\lambda$ to 10 in StyleGAN2, while it is set to 100 in other methods due to StyleGAN2's additional constraint terms. We note a discrepancy between reported scores in the literature and our results using the provided code, possibly due to hardware variations or differences between runs.

\section{More Qualitative Results}

\begin{figure*}[h]
\setlength{\tabcolsep}{1pt}
\centering
\includegraphics[width=\linewidth]{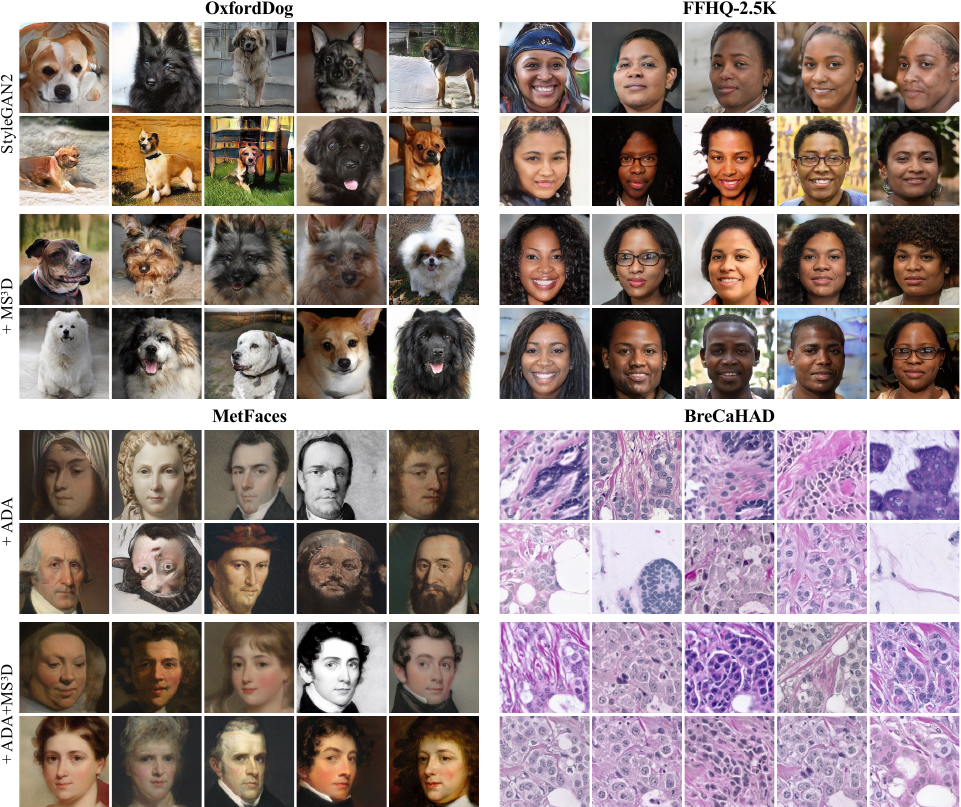}
\caption{More generated images (best FID).}
\label{fig:main_qualitative_full}
\end{figure*}

\begin{figure*}[h]
\setlength{\tabcolsep}{1pt}
\centering
\includegraphics[width=\linewidth]{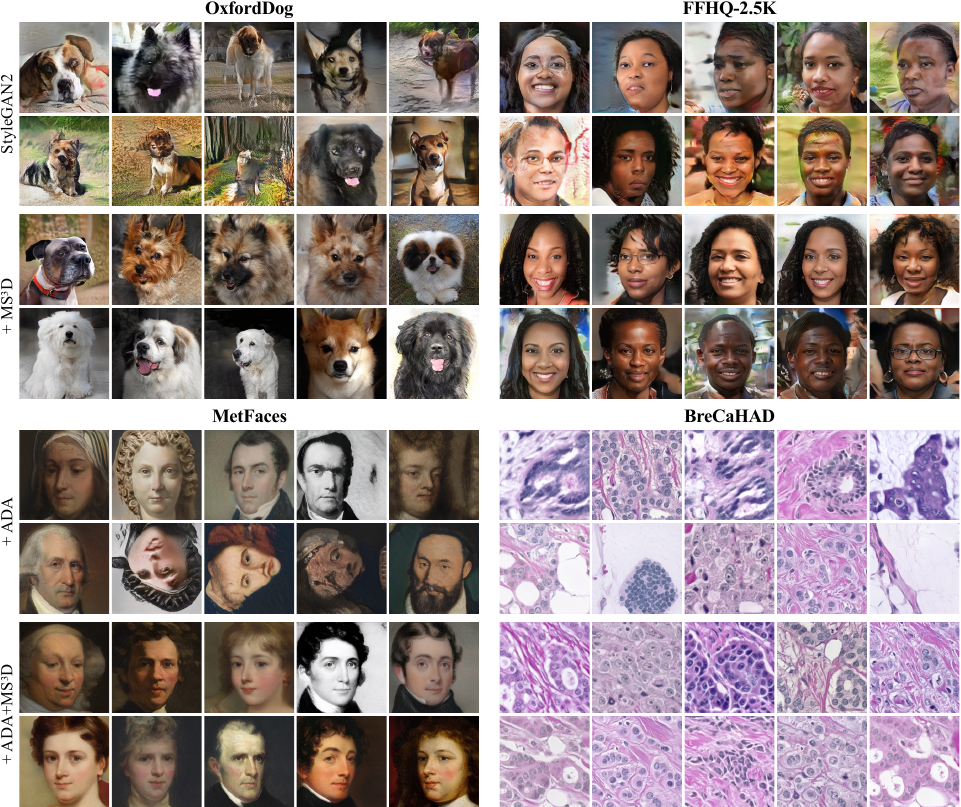}
\caption{More generated images (last iteration).}
\label{fig:main_qualitative_full1}
\end{figure*}

In this section, we present additional qualitative results obtained from OxfordDog, FFHQ-2.5K, MetFaces, and BreCaHAD datasets. We compare the outcomes of StyleGAN2 and StyleGAN2 with MS$^3$D on these datasets, as illustrated in Figs.~\ref{fig:main_qualitative_full} and \ref{fig:main_qualitative_full1}.

\end{document}